\documentclass[sigconf]{acmart}
\AtBeginDocument{%
  \providecommand\BibTeX{{%
    Bib\TeX}}}

\setcopyright{acmlicensed}
\copyrightyear{2018}
\acmYear{2018}
\acmDOI{XXXXXXX.XXXXXXX}
\acmConference[Conference acronym 'XX]{Make sure to enter the correct
  conference title from your rights confirmation email}{June 03--05,
  2018}{Woodstock, NY}
\acmISBN{978-1-4503-XXXX-X/2018/06}

\usepackage{cite}
\usepackage{amsmath,amsfonts}
\usepackage{algorithmic}
\usepackage{graphicx}
\usepackage{textcomp}
\usepackage{comment}
\usepackage{hyperref}
\usepackage{xcolor}
\usepackage{epigraph} 
\usepackage{cleveref}
\usepackage{booktabs}
\usepackage{multirow}
\usepackage{colortbl}
\usepackage[table]{xcolor}
\usepackage[most]{tcolorbox}
\usepackage{xcolor}
\usepackage{enumitem} 
\tcbuselibrary{breakable}
\def\BibTeX{{\rm B\kern-.05em{\sc i\kern-.025em b}\kern-.08em
    T\kern-.1667em\lower.7ex\hbox{E}\kern-.125emX}}
\begin{document}

\definecolor{my-triage-color}{HTML}{96D3D4}
\definecolor{my-triage-light-color}{HTML}{DCEFF3}
\definecolor{my-clinical-rounds-color}{HTML}{6DB9D0}
\definecolor{my-clinical-rounds-light-color}{HTML}{CDE6ED}
\definecolor{my-lab-color}{HTML}{A1AFDA}
\definecolor{my-lab-light-color}{HTML}{DBE0F4}
\definecolor{my-evidence-light-color}{HTML}{E9E1F0}
\definecolor{my-evidence-color}{HTML}{C1A5CE}
\definecolor{my-synthesizer-color}{HTML}{F6D0B1}
\definecolor{my-synthesizer-light-color}{HTML}{FBEBDD}

\newcommand{\pos}[1]{\textcolor{green!60!black}{\textbf{+#1\%}}}
\newcommand{\negat}[1]{\textcolor{red!70!black}{\textbf{-#1\%}}}
\newcommand{\zero}{\textcolor{gray}{0\%}}

\newtcolorbox{triage}{
 colback=my-triage-light-color,   
 colframe=my-triage-color,  
 boxrule=1pt,       
 arc=3pt,         
 left=6pt, right=6pt, top=6pt, bottom=6pt, 
 fonttitle=\bfseries,
 coltitle=black,
}

\newtcolorbox{rounds}{
 colback=my-clinical-rounds-light-color,   
 colframe=my-clinical-rounds-color,  
 boxrule=1pt,       
 arc=3pt,         
 left=6pt, right=6pt, top=6pt, bottom=6pt, 
 fonttitle=\bfseries,
 coltitle=black,
}

\newtcolorbox{labresults}{
 colback=my-lab-light-color,   
 colframe=my-lab-color,  
 boxrule=1pt,       
 arc=3pt,         
 left=6pt, right=6pt, top=6pt, bottom=6pt, 
 fonttitle=\bfseries,
 coltitle=black,
}

\newtcolorbox{evidence}{
 colback=my-evidence-light-color,   
 colframe=my-evidence-color,  
 boxrule=1pt,       
 arc=3pt,         
 left=6pt, right=6pt, top=6pt, bottom=6pt, 
 fonttitle=\bfseries,
 coltitle=black,
}

\newtcolorbox{final}{
 colback=my-synthesizer-light-color,   
 colframe=my-synthesizer-color,  
 boxrule=1pt,       
 arc=3pt,         
 left=6pt, right=6pt, top=6pt, bottom=6pt, 
 fonttitle=\bfseries,
 coltitle=black,
}

\definecolor{promptbg}{RGB}{248,249,250}
\definecolor{promptframe}{RGB}{180,180,180}
\definecolor{varbg}{RGB}{230,235,240}
\definecolor{vartext}{RGB}{0,50,120}

\definecolor{zeroshotcolor}{RGB}{119,119,119}    
\definecolor{triagecolor}{RGB}{149,211,212}      
\definecolor{doctorcolor}{RGB}{109,184,207}      
\definecolor{consultantcolor}{RGB}{25,39,66}     
\definecolor{codercolor}{RGB}{128,139,173}       
\definecolor{synthesizercolor}{RGB}{246,207,177} 
\definecolor{evidencecolor}{RGB}{193,165,205}    

\newcommand{\pyvar}[1]{%
  \colorbox{varbg}{\textcolor{vartext}{\texttt{\small #1}}}%
}

\newcommand{\prompthead}[1]{%
  \par                
  \addvspace{1.4em}   
  \noindent           
  \textbf{\large #1}  
  \par                
  \vspace{0.2em}      
  \nopagebreak        
}

\newtcolorbox{promptbox}[2][zeroshotcolor]{
  colback=promptbg,
  colframe=#1,
  coltitle=white,
  colbacktitle=#1,
  boxrule=0.3mm,
  title={\textbf{#2}},
  fontupper=\small\sffamily, 
  sharp corners=south,
  arc=3mm,
  parbox=false, 
  breakable     
}

\title{
A Multi-Agent Framework for Interpreting Multivariate Physiological Time Series
}

\author{Davide Gabrielli}
\orcid{0009-0000-5672-6749}
\affiliation{%
\institution{Sapienza University of Rome}
\city{Rome}
\country{Italy}
}

\author{Paola Velardi}
\orcid{0000-0003-0884-1499}
\affiliation{%
\institution{ISTC-CNR \& Sapienza University of Rome}
\city{Rome}
\country{Italy}
}

\author{Stefano Faralli}
\orcid{0000-0003-3684-8815}
\affiliation{%
\institution{Sapienza University of Rome}
\city{Rome}
\country{Italy}
}

\author{Bardh Prenkaj}
\authornote{Corresponding: bardhprenkaj95@gmail.com}
\orcid{0000-0002-2991-2279}
\affiliation{%
 \institution{Technical University of Munich}
 \city{Munich}
 \country{Germany}}

\begin{abstract}
Continuous physiological monitoring is central to emergency care, yet deploying trustworthy AI is challenging. While LLMs can translate complex physiological signals into clinical narratives, it is unclear how agentic systems perform relative to zero-shot inference. To address these questions, we present Vivaldi, a role-structured multi-agent system that explains multivariate physiological time series. Due to regulatory constraints that preclude live deployment, we instantiate Vivaldi in a controlled, clinical pilot to a small, highly qualified cohort of emergency medicine experts, whose evaluations reveal a context-dependent picture that contrasts with prevailing assumptions that agentic reasoning uniformly improves performance. Our experiments show that agentic pipelines substantially benefit non-thinking and medically fine-tuned models, improving expert-rated explanation justification and relevance by +6.9 and +9.7 points, respectively. Contrarily, for thinking models, agentic orchestration often degrades explanation quality, including a 14-point drop in relevance, while improving diagnostic precision (ESI F1 +3.6). We also find that explicit tool-based computation is decisive for codifiable clinical metrics, whereas subjective targets, such as pain scores and length of stay, show limited or inconsistent changes. Expert evaluation further indicates that gains in clinical utility depend on visualization conventions, with medically specialized models achieving the most favorable trade-offs between utility and clarity. Together, these findings show that the value of agentic AI lies in the selective externalization of computation and structure rather than in maximal reasoning complexity, and highlight concrete design trade-offs and learned lessons, broadly applicable to explainable AI in safety-critical healthcare settings.
\end{abstract}
\maketitle

\begin{center}
\centering
\textit{"Agentic AI is best understood as an ensemble that benefits specialized, smaller models, rather than as a general-purpose performance multiplier."}
\end{center}

\section{Introduction}
Deploying AI systems in safety-critical domains such as emergency medicine remains a central challenge for applied data science, especially in the EU due to stricter regulations \citep{van2024eu}. In healthcare, models often perform well on retrospective benchmarks, but fail to deliver real-world value when clinicians cannot quickly understand, trust, and act on their outputs. Hence, effective deployment requires predictive accuracy,  explainability, clinical plausibility, and alignment with real decision-making workflows \citep{Sadeghi2024Review,Rosenbacke2024ClinicianTrust,context_xai_2025}. These limitations are especially acute for multivariate physiological time series, where clinicians must interpret alerts relative to patient-specific baselines, uncertainty, and clinical context. Although unsupervised and personalized anomaly detection enables continuous monitoring without exhaustive labeling \citep{Gabrielli2025AIOnThePulse}, anomaly scores alone do not support clinical decision-making. \textit{Clinicians need explanations that link observed deviations to recognizable physiological patterns, contextual drivers, and actionable next steps} \citep{Gabrielli2025AIOnThePulse,context_xai_2025}.

Recent work proposes using LLMs to translate complex physiological signals into natural-language explanations \citep{tsaia_2026,zheng2025promind}. However, deploying LLM-based systems in clinical settings raises unresolved challenges that offline evaluation alone cannot address.
 In particular, three challenges remain underexplored:

\noindent\textbf{(C1) Structuring explanations for clinical workflows.}
Explanations must reflect how clinicians reason in practice, integrating safety screening, hypothesis formation, critique, and evidence synthesis under time and attention constraints \citep{brambilla2022analysis,graber2004structure}.

\noindent\textbf{(C2) Understanding when agentic reasoning helps.}
While agentic and tool-augmented LLM pipelines are increasingly proposed, it is unclear whether externally decomposed reasoning consistently improves interpretability, relevance, and trust compared to simpler zero-shot inference, especially for bigger models \citep{schick2023toolformerlanguagemodelsteach,yao2023treethoughtsdeliberateproblem,yao2023reactsynergizingreasoningacting}.

\noindent\textbf{(C3) Aligning explanations with clinician expectations.}
Human-centered XAI shows that explanation utility depends on the recipient's role and expertise \citep{Chromik2021HumanCentric,Dey2022HealthcarePersona,Kim2024StakeholderTailored}. For medical experts, explanations must be concise, context-aware, and grounded in familiar clinical representations; overly verbose or unconventional outputs can reduce trust and adoption \citep{Rosenbacke2024ClinicianTrust,Sadeghi2024Review}.

In this work, we present Vivaldi, a role-structured multi-agent system for interpreting multivariate physiological time series in emergency department (ED) settings. Vivaldi instantiates explanation generation as a multi-agent \emph{clinical team}, mirroring real ED workflows: (1) a triage agent computes safety metrics and personalized thresholds; (2) a doctor iteratively forms and revises hypotheses from longitudinal vital-sign trends; (3) a consultant critiques and fine-tunes the doctor's hypotheses; (4) a coder performs quantitative analyses and generates visual evidence; and (5) a synthesizer prepares the final explanation. An orchestrator maintains shared clinical state, coordinates agent interactions, and manages evidence selection to reflect real-world constraints.
 
This paper makes the following contributions, each addressing one or more of the challenges outlined above:

\noindent\textbf{(1) Controlled clinical evaluation of agentic explanations (C1--C3).} We evaluate Vivaldi through a controlled clinical pilot conducted under regulatory constraints, in which emergency medicine domain experts used the system to review patient-derived physiological data. This deployment yielded 109 anonymous expert evaluations, providing quantified post-deployment evidence across six clinically grounded dimensions (i.e., factuality, relevance, justification, trust, visualization clarity, and decision utility). Given the specialized nature of the domain, this corpus of expert feedback ensures thematic saturation and provides a basis for assessing the system's clinical alignment.

\noindent\textbf{(2) Model-dependent effects of agentic reasoning on explanation quality (C2).}   We present a systematic, within-subject comparison between agentic and zero-shot inference across five LLM families, showing that agentic decomposition improves expert-rated relevance and justification for non-thinking and medically specialized models ($+6.9$ and $+9.7$ points, respectively), while often degrading relevance and trust (up to $-14$ points) for thinking models with strong zero-shot baselines.

\noindent\textbf{(3) Quantified benefits of explicit tool-based computation for triage metrics (C1, C2).} Delegating numerical reasoning to executable tools improves performance from an F1 score of $61.0$ to $64.6$ (thinking models) and $40.7$ to $65.4$ (non-thinking models), while yielding perfect or near-perfect performance on codifiable metrics (\autoref{tab:triage_metrics}). In contrast, performance on subjective outcomes (i.e., pain score, length of stay) shows no consistent improvement.

\noindent\textbf{(4) Clinician-grounded analysis of visualization trade-offs (C3).}   Using expert ratings, we quantify how agentic workflows affect clinical utility and chart comprehensibility, showing consistent gains in perceived utility across models but model-dependent and often small or negative changes in visualization comprehensibility.

\noindent\textbf{(5) Clarifying when agentic AI helps, and when it hurts (C2).} We show that agentic decomposition is \emph{not} a universal improvement. While it often degrades relevance and trust in bigger models, it consistently benefits smaller, medically fine-tuned models, where explicit role specialization and external computation compensate for limited internal reasoning.

\section{Related Work}

The landscape of physiological monitoring is moving from task-specific supervised models toward more general architectures that can model complex physiology across heterogeneous sensors. We organize prior work into four main streams.

\noindent\textbf{Universal Temporal Foundation Models and Generalist Architectures.}
Universal time-series foundation models detach from the ``one model, one task'' paradigm. Architectures such as UniTS \citep{gao2025units} unify forecasting, classification, and anomaly detection within a single backbone via task tokenization, outperforming specialized models like TimesNet \citep{wu2022timesnet}. Large-scale pretraining has enabled models such as Moirai \citep{moirai_2024} and TimesFM \citep{das2024decoder} to learn general temporal priors, supporting robust zero-shot transfer in healthcare settings with limited labeled data. Similarly, TimeGPT demonstrates that transformer-based forecasting can achieve strong performance in healthcare operations with minimal task-specific adaptation \citep{garza2023timegpt}.

\noindent\textbf{Behavior-Aware Monitoring and Causal Reasoning}
This line of work emphasizes behavior-aware representations that align better with clinically meaningful timescales than raw high-frequency signals. Large-scale studies show that behavioral features capture physiologically relevant structure more effectively \citep{wbm_2025}. To support efficiency on edge devices, research has increasingly adopted Structured State Space Models (SSMs), which scale linearly with sequence length \citep{somvanshi2025s4}. Compact SSMs, such as ECG-CPC, can outperform transformer models orders of magnitude larger \citep{al2025benchmarking}. To connect statistical anomalies with clinical interpretation, approaches like ProMind-LLM integrate sensor data with contextual records using causal Chain-of-Thought reasoning \citep{zheng2025promind}.

\noindent\textbf{Human-Centered and Persona-Aware XAI.} Recent XAI roadmaps identify context- and user-dependent explanations as essential for clinical adoption \citep{context_xai_2025}. For clinicians, explanations must be grounded in pathophysiology and support decision-making, whereas for patients, they aim to improve health literacy and engagement \citep{moell2025journaling,pal2025generative}. Evaluation has shifted toward multi-level benchmarks such as TSAIA, which test evidence integration and multi-step reasoning \citep{tsaia_2026}, and expert-driven frameworks like CLEVER, which emphasize clinical relevance and factual grounding \citep{kocaman2025clinical}. Hybrid personalization strategies combining population-level models with individual N-of-1 adaptation have been proposed to address generalizability and privacy trade-offs \citep{konigorski2026personalization}.

\noindent\textbf{Multi-Agent Clinical Systems and Tool-Augmented LLMs.}
Recent work increasingly models clinical reasoning as coordination among specialized agents rather than a single monolithic model. Multi-agent frameworks such as MACD \citep{li2025macd} and MedLA \citep{ma2025medla} improve diagnostic accuracy and traceability through role decomposition, while collaborative agent systems have been applied to clinical note analysis \citep{lee2025automatedclinicalproblemdetection}. These approaches show that agentic decomposition can enhance robustness, but they primarily evaluate diagnostic correctness rather than interactive interpretability or visualization usability. Complementary work on tool-augmented reasoning demonstrates that delegating deterministic computation to external tools improves numerical reliability (Toolformer \citep{schick2023toolformerlanguagemodelsteach}), while frameworks such as ReAct \citep{yao2023reactsynergizingreasoningacting} and Tree-of-Thought \citep{yao2023treethoughtsdeliberateproblem} structure multi-step reasoning as explicit action or search processes.\newline
\newline
\noindent\textbf{Our contribution.}
Our pipeline builds on these insights by (i) deterministically computing safety-critical indices (Shock Index, qSOFA, MAP) in a sandboxed coder agent, and (ii) using role-conditioned LLM agents to generate and select clinically meaningful visual evidence, a combination not directly evaluated by prior multi-agent medical systems.
A recent systematic review of multi-agent CDSSs~\citep{SILVEIRA2026114447} identifies two persistent gaps addressed by our work: (a) the lack of rigorous, clinician-centered evaluation of explainability, and (b) insufficient mechanisms for integrating and coordinating heterogeneous clinical evidence in ways that align with real-world clinical workflows. Our anonymous expert evaluation of agentic versus zero-shot pipelines provides direct empirical evidence on these questions and complements the diagnostic accuracy focus of prior multi-agent systems.
\begin{figure}[!t]
    \centering
    \includegraphics[width=\linewidth]{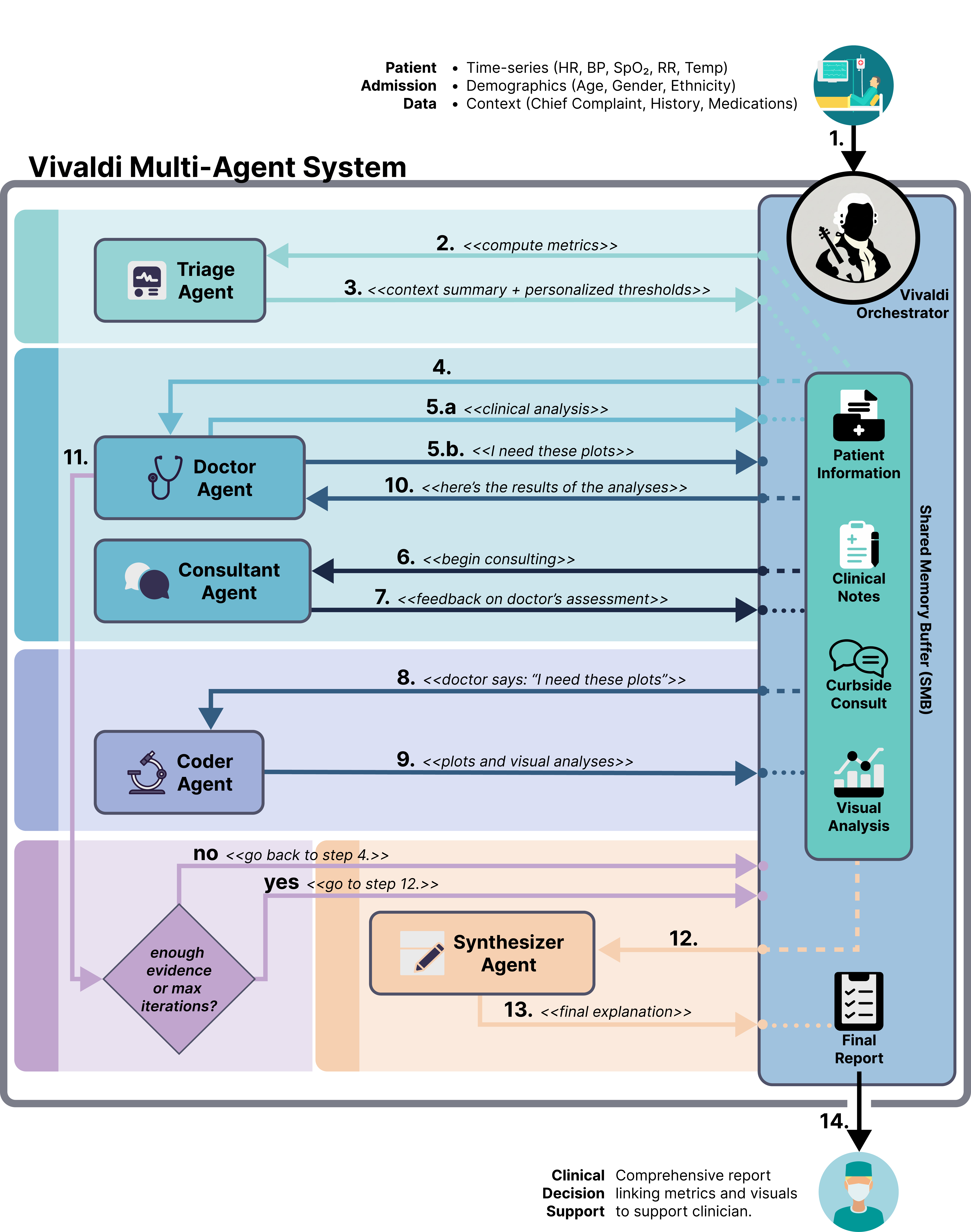}
    \caption{System architecture as a five-scene Emergency Department narrative. The area on the right represents the Shared Memory Buffer (SMB) that Vivaldi manages, orchestrating communication with it and the agents, or between agents, as an intermediary: e.g., think of this as API calls where read (dashed lines) and write (dotted lines) operations happen in the SMB. The circles at the edge of the shaded area on the right illustrate the agents' entry point for communicating with Vivaldi. We remove the details of read/write interactions between the SMB and Vivaldi for illustration purposes, and defer them to the single scenes. Same-colored lines represent a single logical flow whose completion might depend on multiple interactions between Vivaldi and the agents therein (e.g., 5.b, 8, 9, and 10).} 
    \label{fig:architecture}
\end{figure}
\section{Dataset and Preprocessing}
\label{sec:dataset}

We use the Multimodal Clinical Monitoring in the Emergency Department (MC-MED) dataset \citep{kansal2025multimodal, kansal2025mc, PhysioNet}, a large-scale, de-identified collection of emergency department (ED) visits from an academic medical center. While MC-MED contains diverse modalities including waveforms and free text, our analysis focuses specifically on numeric vital signs, demographics, patient history, and medications:

\noindent\textbf{(1) Visit Logic and Demographics:} The file \texttt{visits.csv} is the core index for patient encounters. We extracted demographic variables (Age, Gender, Ethnicity). We capture the clinical context via the Chief Complaint field, while we choose the Emergency Severity Index (ESI) and Length of Stay (LOS) as ground-truth targets.

\noindent\textbf{(2) Vital Signs:} We use physiologic data from \texttt{numerics.csv}, categorizing them into core vitals (Heart Rate, Systolic and Diastolic Blood Pressure, Oxygen Saturation) and medium-frequency vitals (Respiratory Rate, Temperature). We extract pain scores as the ground truth target.

\noindent\textbf{(3) Patient History:} Past medical history was extracted from \texttt{pmh.csv} to establish baseline comorbidities.

\noindent\textbf{(4) Medications:} Active home medications were identified from \texttt{meds.csv}, prioritizing the ``Generic Name'' for standardization.
\newline\newline
\noindent\textbf{Preprocessing.}
To ensure data quality, we implemented a standardized preprocessing pipeline captured in a static benchmarking index. This index enforces rigorous inclusion criteria, selecting visits only if they have valid ground-truth labels (integer ESI, non-null LOS, and valid pain scores) and sufficient data density (minimum of 30 samples for both core and medium-frequency vitals within a continuous 24-hour window). The index persists precise temporal boundaries to guarantee that all models evaluate identical temporal slices. During runtime, features are automatically normalized (e.g., unit standardization to Celsius) and cleaned of artifacts using physiological plausibility filters (rejecting values outside clinically valid ranges, e.g., HR 30--220 bpm, SpO$_2$ 70--100\%, SBP 50--250 mmHg). We make this benchmark index publicly available to facilitate consistent evaluation and reproducibility in future research.

\section{Method}\label{sec:method}

We frame our method as a realistic simplification of how an emergency department (ED) team manages an acute case \citep{graber2004structure,brambilla2022analysis}. Consider a senior patient arriving with dyspnea and chest pressure. In practice, the workup unfolds as a structured relay: (1) triage establishes safety baselines; (2) the attending forms and revises hypotheses; (3) specialists stress-test the plan; (4) quantitative analyses are computed on demand; (5) and a senior clinician synthesizes the case into a final disposition. Our system instantiates this choreography as a multi-agent \emph{clinical team}. Each clinical role is embodied by a specialized agent, while an orchestrator (that we name  Vivaldi\footnote{\url{https://anonymous.4open.science/r/vivaldi-xai-ed/}}) maintains shared clinical state, manages handoffs, and controls iteration. We invite the reader to cross-check between the system architecture (\Cref{fig:architecture}) and the details of each scene. In each scene, \textbf{R}/\textbf{W} are read/write operations from and to the SMB.

\begin{triage}
\textbf{Scene 1: Triage.}  \textit{The story begins at triage. A nurse rapidly captures vitals, computes quick screening scores, and adjusts what ``normal'' means based on the patient's history.}\newline\newline
\includegraphics[width=\linewidth]{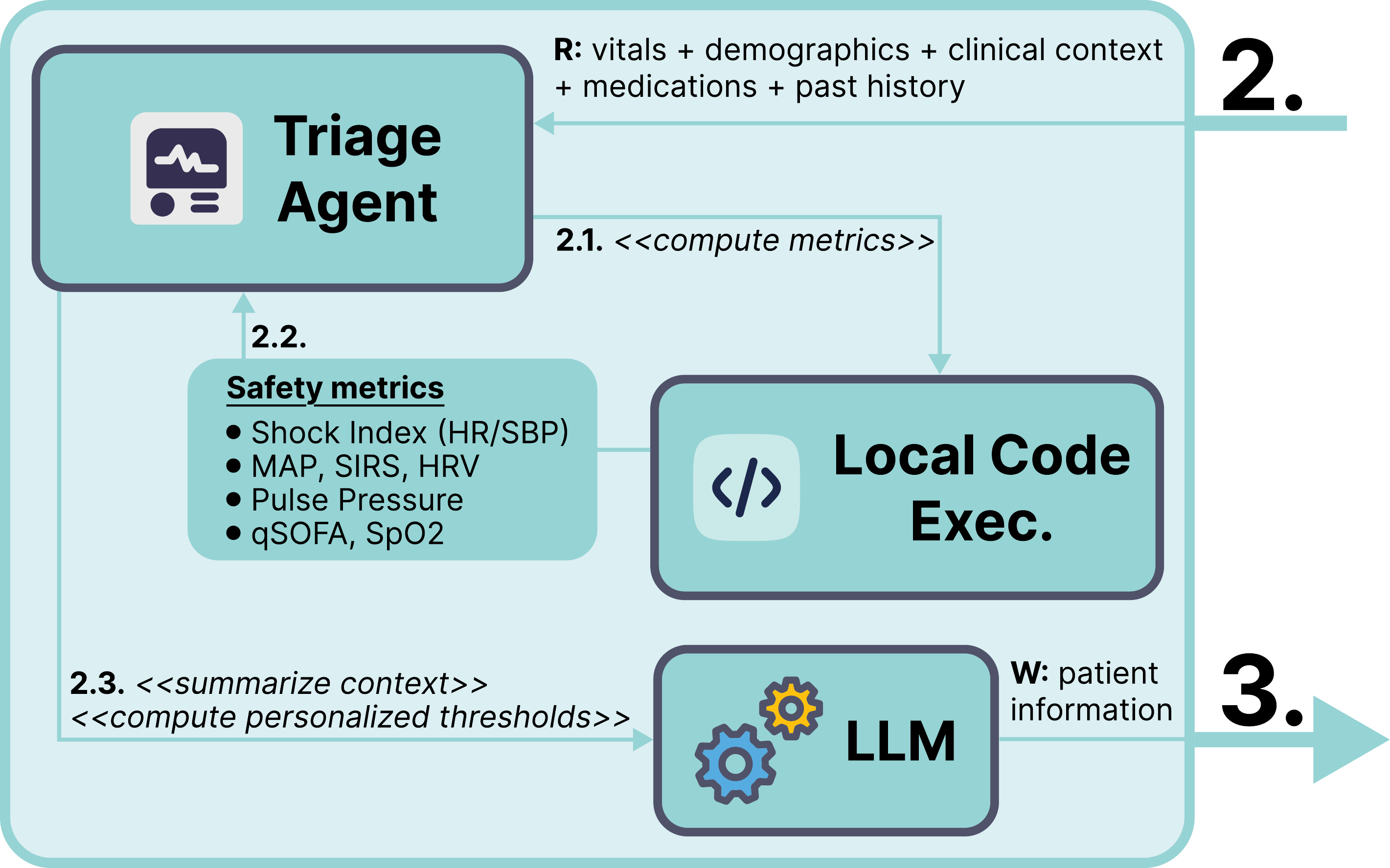}
\end{triage}
\noindent When Vivaldi receives a patient (step 1), it delegates the computation to \texttt{TriageAgent} (step 2), which performs a three-step triaging. First, it computes deterministic safety metrics (step 2.1) directly from raw vital-sign time series
using local code (not an LLM) to ensure numerical precision and low latency (step 2.2). Second, it invokes a role-conditioned LLM (step 2.3) to translate past medical history and medications into a structured clinical context summary and personalized threshold recommendations. For example, it may adjust blood pressure warning bands for chronic hypertension, or revise fever interpretation if the profile suggests axillary measurement. The agent then renders a multi-panel vitals visualization with individualized threshold bands overlaid (step 3). These triage outputs\footnote{We store (i) a dictionary of safety metrics, (ii) a narrative context string explaining personalized thresholds, and (iii) a vitals panel image.} are then stored by Vivaldi into the patient information data structure within the Shared Memory Buffer (SMB).

\begin{rounds}
\textbf{Scene 2: Clinical Rounds.}  \textit{Next, the attending physician reviews the chart and begins the diagnostic workup, typically iterating as new information clarifies risk and trajectory.}\newline\newline
\includegraphics[width=\linewidth]{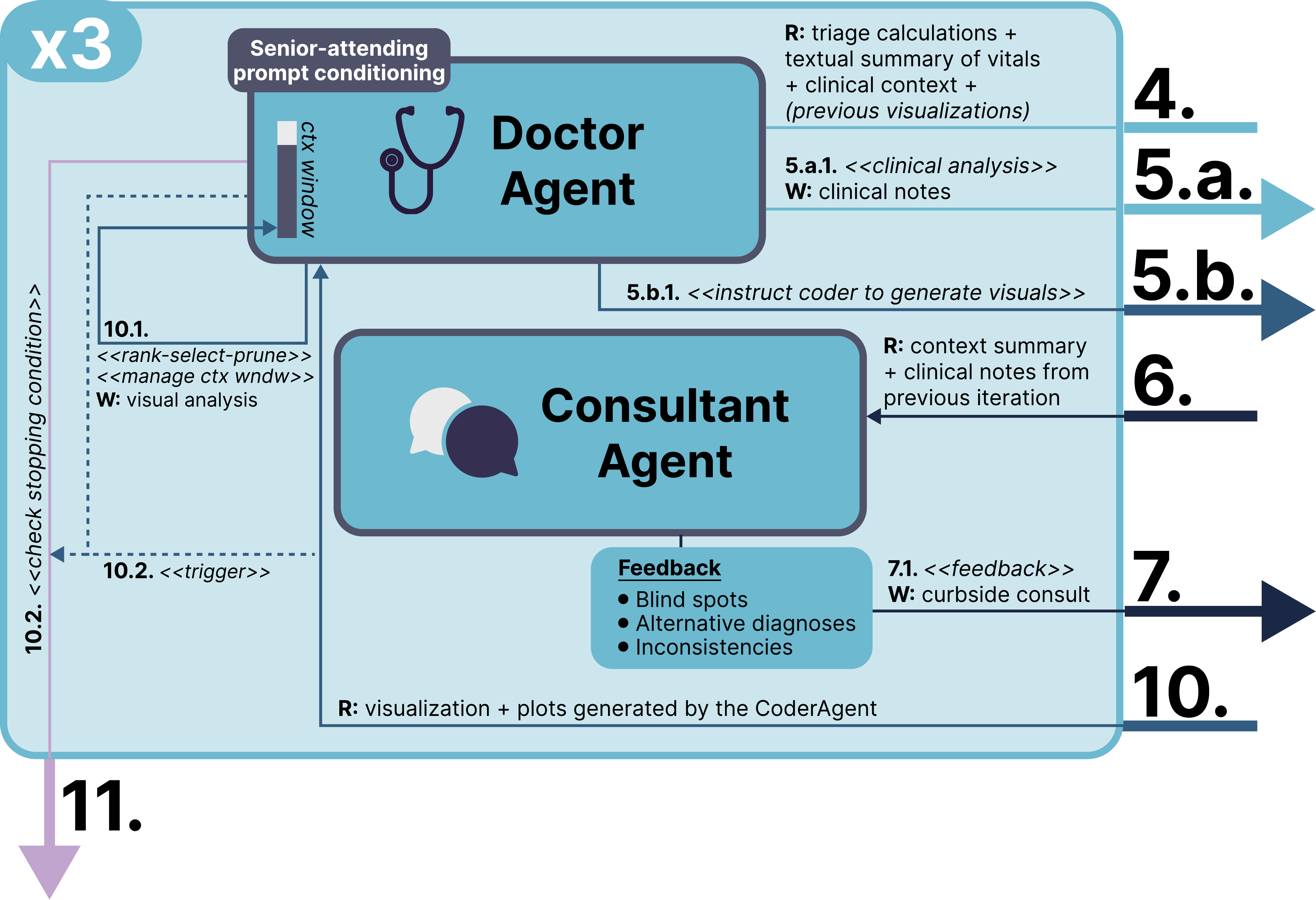}
\end{rounds}
\noindent Our \texttt{DoctorAgent} mirrors this behavior through an iterative rounds loop (steps 4, 5.a, 5.b).\footnote{There are up to 3 iterations. The \texttt{DoctorAgent} can terminate the rounds early if it finds the discussion sufficient.} At the start of each round, Vivaldi presents (step 4) the doctor with: (i) the current calculation results from triage and prior computations, (ii) a textual summary of the vital-sign time series, (iii) the triage-derived clinical context, and (iv) previously reviewed visualizations (the triage panel and any prior plots). Conditioned on a senior-attending prompt, the doctor produces a free-text clinical analysis with evolving hypotheses, concerns, and a provisional plan. These analyses are appended to the Doctor's history window, forming a longitudinal narrative analogous to progress notes. To manage context limits, the Doctor employs a \textit{rank-select-prune} strategy: it assigns a clinical relevance score (1--10) to each new plot, retains only the top-3 visualizations (removing the others from the context), and terminates the loop early if the cumulative evidence is deemed definitive.

\textit{ED reasoning is rarely solo.} In emergency practice, clinicians frequently rely on informal ``curbside consultations,'' in which a colleague provides a second opinion.
After each analysis (step 5.a), Vivaldi calls a \texttt{ConsultantAgent} (step 6). The consultant receives the same evidence bundle and the doctor's analysis of the previous iteration, and returns a critique that (i) surfaces blind spots, (ii) proposes alternative diagnoses, and (iii) flags inconsistencies (step 7). This feedback is useful for the final explanation summarization (steps 12 and 13).

\begin{labresults}
\textbf{Scene 3: Lab Results.} \textit{ED workups depend on computations that are not obvious from raw plots, such as cross-correlations, trend alerts, and pattern summaries.}\newline\newline
\includegraphics[width=\linewidth]{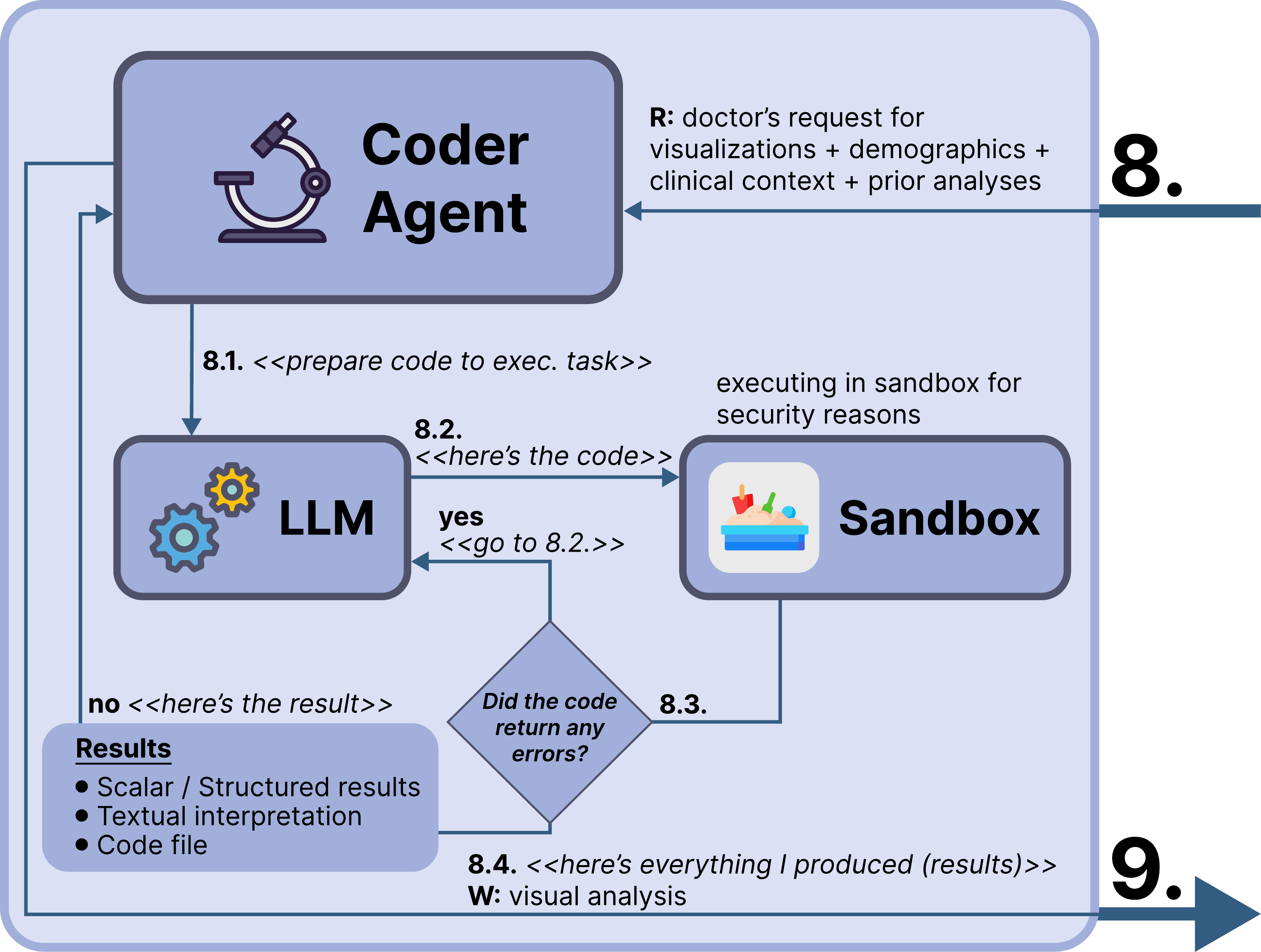}
\end{labresults}
In each clinical round, the \texttt{DoctorAgent} performs a second structured pass (step 6) that prescribes concrete computational tasks (e.g., \textit{``compute MAP trajectory over the last 90 minutes and flag periods below 65\,mmHg''}). Vivaldi forwards these task commands, together with vitals, demographics, clinical context, and all prior results, to the \texttt{CoderAgent} (step 8). This agent uses an LLM to write Python code for each requested analysis (steps 8.1, 8.2, 8.3). Each successful execution produces (i) a scalar or structured result, (ii) a short interpretation string, (iii) paths to generated figures, and (iv) the executed code for auditability. Vivaldi adds these outputs to the shared case state (step 9), making them available to the other agents for the next iterations. 

\begin{evidence}
\textbf{Scene 4: Selecting Evidence.} \textit{Clinicians do not review every plot; they select a small set of high-yield visualizations that best justify a decision.}\newline\newline
\includegraphics[width=\linewidth]{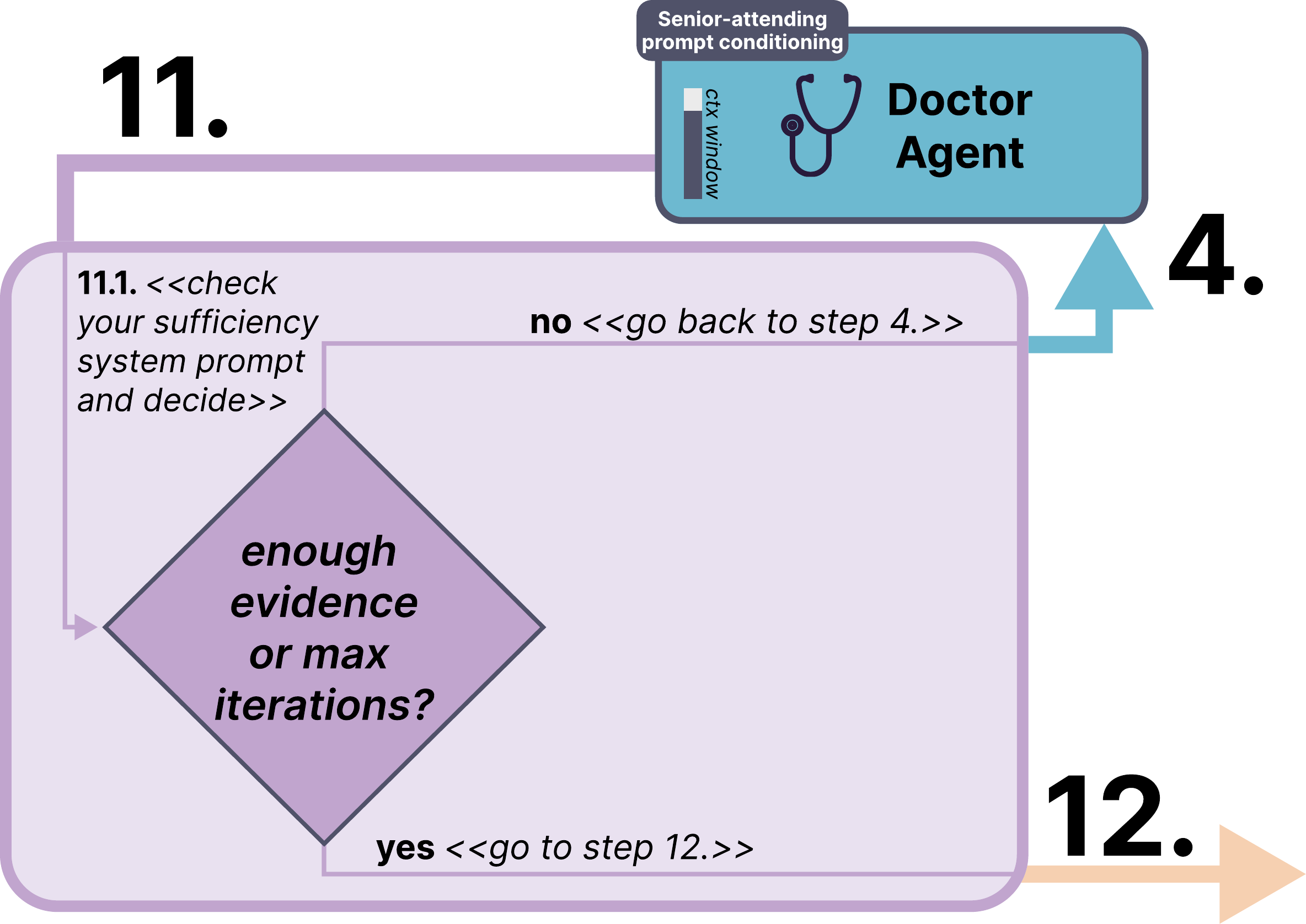}
\end{evidence}
Notice that this scene is closely related to the second one, so we invite the reader to view it in relation to it. 

After each batch of \texttt{CoderAgent} outputs, Vivaldi returns to the \texttt{DoctorAgent} with only the new figures, along with the current clinical context and the prior analysis (step 10). The doctor reviews these images multimodally, ranks them by relevance (see step 10.1 in Scene 2), and writes short captions explaining why each figure matters. Vivaldi merges these rankings with any existing image shortlist and retains only the top-\(N\) plots (we simplified this interaction into a single step in 10.1). In parallel, the doctor emits a sufficiency flag (step 10.2), which is triggered by the visualizations generated by the \texttt{CoderAgent}. If it judges that the current evidence is adequate for safe disposition (see \textbf{yes} in step 11.1), Vivaldi stops clinical rounds and passes the current findings for final synthesis. Otherwise (see \textbf{no} in step 11.1), Vivaldi repeats the rounds (step 4).

\begin{final}
\textbf{Scene 5: Final Synthesis.} \textit{Finally, a senior attending reviews the entire workup, integrates all evidence, and commits to an assessment and disposition.}\newline\newline
\includegraphics[width=\linewidth]{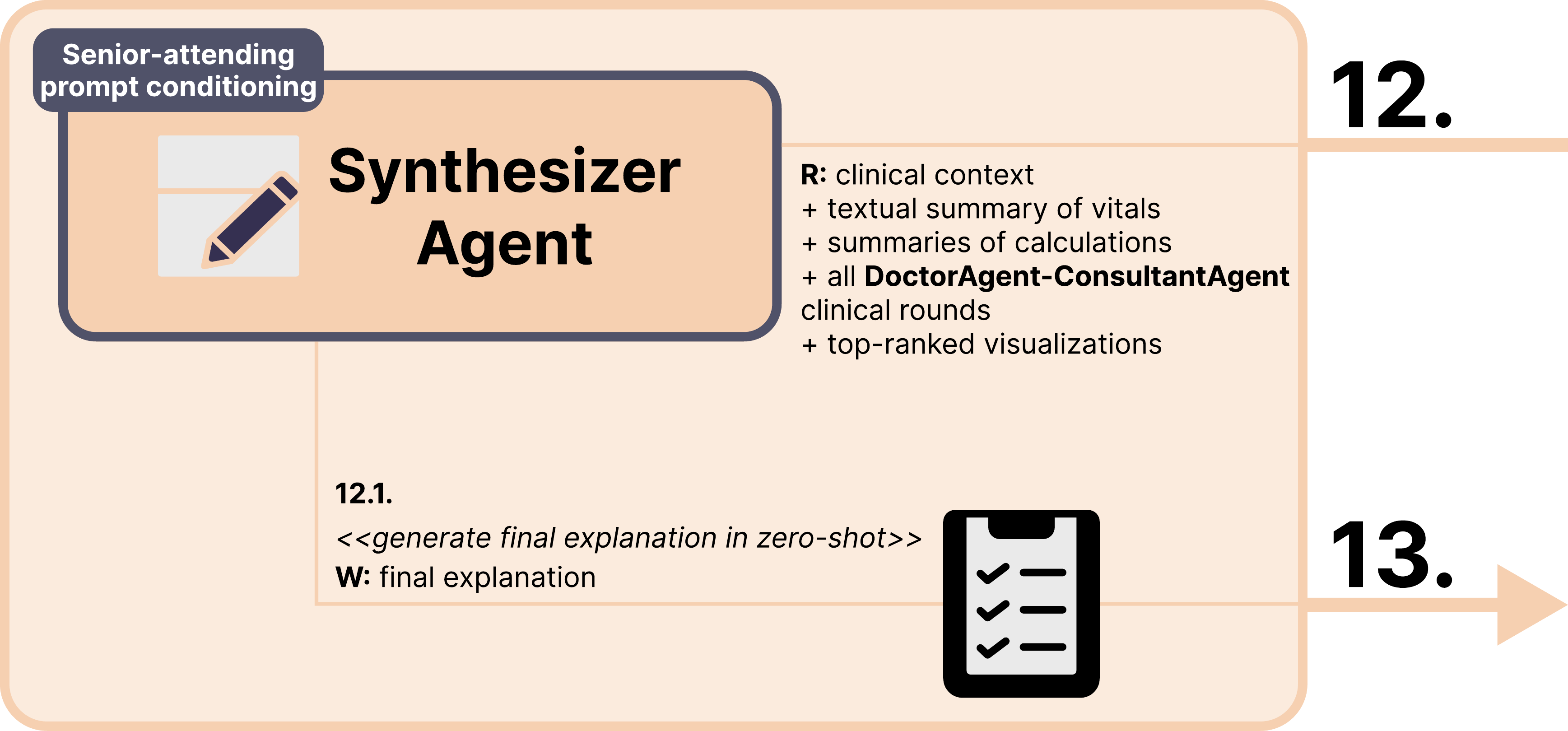}
\end{final}
\noindent Once the case is deemed sufficient (or the maximum iterations are reached), Vivaldi constructs a final evidence bundle: (i) the triage clinical context, (ii) a vitals summary, (iii) concise summaries of all successful calculations, (iv) the evolution of \texttt{DoctorAgent}--\texttt{ConsultantAgent} narratives across iterations, and (v) descriptions of the top-ranked figures (12). Conditioned on a senior-attending prompt and a structured output schema, the \texttt{SynthesizerAgent} produces the final assessment in zero-shot mode (steps 12.1 and 13). Critically, this agent is instructed to preserve the doctor's decisions unless there is an obvious inconsistency, reflecting real ED dynamics where a senior attending validates and formalizes a junior team's plan rather than re-deriving it from scratch.

\section{Use-case Setup}

\textbf{Experimental Setup and Baselines.} Our architecture, built as a combination of custom modular classes and LangChain~\citep{langchain} primitives, permits us to benchmark five distinct model families: GPT 5.2, Claude 4.5 Opus, Google Gemini 3 Pro, Llama 4 Maverick, and MedGemma 27B. By keeping the prompt logic and orchestration layer constant across models, we isolate the effect of the underlying reasoning engine on agentic performance.
We compare two distinct inference strategies. First, the \textbf{Zero-Shot} approach serves as a baseline, representing the current standard for LLM interaction. Here, a single model instance receives the full patient context -- including demographics, past medical history, medications, and vital signs -- in a single prompt. We instruct the model to generate a complete clinical assessment, compute necessary metrics, and determine the final disposition without intermediate reasoning steps or the use of external tools. Second, the \textbf{Agentic} approach instantiates the multi-agent architecture detailed in \Cref{sec:method}, decomposing the inference process into specialized roles rather than relying on a single monolithic context window.\newline
\newline
\noindent\textbf{Implementation and Reproducibility.}
To ensure reproducibility, we enforce strict constraints on agent configuration and orchestration (see \autoref{sec:appendix_prompts} for full system prompts). We initialize the \texttt{DoctorAgent} with a ``Senior Attending'' persona, explicitly instructed to review previous estimates and update assessments based on new data. To simulate realistic peer review, we adopt a ``Curbside Consult'' persona with instructions to identify clinical gaps and explicitly request visualizations rather than merely suggesting rule-outs for the \texttt{ConsultantAgent}. We constrained the \texttt{SynthesizerAgent} to produce a coherent narrative and adhere to the \texttt{DoctorAgent}'s final decisions for structured targets such as ESI level, pain score, and length of stay.
To ensure data quality, we employed the standardized preprocessing pipeline and strict inclusion criteria detailed in \autoref{sec:dataset}, including physiological plausibility filters and minimum data density thresholds.
We capped clinical rounds at 3 iterations. The sufficiency of evidence was determined by the \texttt{DoctorAgent} as described in \autoref{sec:method}, enforced via specific instructions in the system prompt.\newline
\newline
\noindent\textbf{Study Design and Participants.}
We conducted an anonymous expert evaluation to assess the clinical utility and interpretability of AI-generated explanations. The study included six clinical experts, primarily from Emergency Medicine and Internal Medicine, who reviewed model outputs associated with real patient cases.  
Each expert evaluated explanations generated by the zero-shot and agentic variants for the same clinical case. In total, the evaluation yielded 109 valid expert reviews. While recruiting highly specialized domain experts is logistically challenging, our cohort size aligns with early-stage clinical evaluation guidelines, which prioritize the depth of formative assessment over large-scale statistical power \citep{Vaseye070904}. Consistent with the qualitative framework of ``informational power'' \citep{malterud2016sample}, we posit that the high specificity of our experts (board-certified physicians) and the dense, structured nature of the reviews provide sufficient signals to identify core systemic issues.\newline
\newline
\noindent\textbf{Evaluation Protocol.}
For each case, we provide experts with standardized clinical inputs, including triage information, vital signs, time-series visualizations, and the corresponding AI-generated explanation. To ensure a rigorous assessment that captures both linguistic quality and multimodal reasoning, we structure our evaluation along six dimensions. We adapt the core metrics of \textit{Factuality} and \textit{Clinical Relevance} from \citep{kocaman2025clinical}, while extending the protocol with specific metrics to assess visualization quality, evidence grounding, and safety-critical reliability:

\noindent\textbf{(1) Factuality} evaluates whether reported clinical facts and underlying mechanisms are correct, serving as a check against inaccuracies (\textit{No / Partially / Yes}).
    
\noindent\textbf{(2) Justification} checks if the conclusions are supported by the selected data and charts (\textit{No / Partially / Yes}).
    
\noindent\textbf{(3) Relevance} evaluates whether the information applies to the specific patient case rather than generic medical knowledge (\textit{No / Partially / Yes}).
    
\noindent\textbf{(4) Trust} evaluates whether the provided explanation is sufficiently robust and safe to be integrated into clinical reasoning. It acts as a safety filter to prevent medical harm, ensuring the logic is dependable regardless of how fluently it is presented (\textit{No / Partially / Yes}).
    
\noindent\textbf{(5) Chart Comprehensibility} measures the ease of interpreting visualizations and their alignment with familiar clinical conventions (5-point Likert scale: 1=\textit{Very difficult} to 5=\textit{Extremely clear}).
    
\noindent\textbf{(6) Clinical Utility} rates how well the explanation aids decision-making and offers practical value in emergency workflows (5-point Likert scale: 1=\textit{Useless} to 5=\textit{Decisive}).

\section{Experiments and Discussion}\label{sec:experiments}
We present a comparative analysis of agentic versus zero-shot inference, disentangling the effects of role-based orchestration from the inherent capabilities of the underlying base models. To rigorously assess where agentic decomposition adds value versus where it introduces overhead, we structure our findings around three research questions. These specifically isolate the impact of the Vivaldi framework on textual explanation quality (RQ1), the precision of tool-delegated triage metrics (RQ2), and the trade-offs between clinical utility and visualization conventions (RQ3).\newline
\newline
\noindent\textbf{RQ1: How does agentic inference affect the clinical trustworthiness, relevance, justification, and trustworthiness of LLM-generated explanations compared to zero-shot reasoning?}
We evaluate explanation quality in \Cref{tab:thinking_vs_nonthinking,tab:model-delta-multirow}. Overall, agentic pipelines do not uniformly outperform zero-shot baselines. Instead, they induce systematic and model-dependent trade-offs that are clinically meaningful. When results are aggregated by model class (\Cref{tab:thinking_vs_nonthinking}), non-thinking models benefit substantially from agentic execution, exhibiting clear gains across all dimensions, particularly relevance (+10.4 points), justification (+7.3 points), and factuality (+4.0 points). Contrarily, thinking-enabled models experience consistent degradations under agentic decomposition, with drops in relevance (-14.5), justification (-9.9), trust (-7.6), and factuality (-5.3). This divergence suggests that models with internal reasoning may already perform implicit multi-step abstraction, and that introducing external agentic structure can diffuse attention, encourage over-generation, or reduce focus on the most salient clinical factors.

\Cref{tab:model-delta-multirow} examines the differences between thinking and non-thinking models. Models like Gemini 3 Pro and Claude 4.5 Opus show significant declines under agentic orchestration, with Gemini 3 Pro experiencing notable drops in relevance (-17.2\%) and justification (-12.6\%), while Claude 4.5 Opus drops across all metrics, particularly relevance (-11.2\%). This suggests agentic pipelines can disrupt the reasoning dynamics of advanced proprietary models. In contrast, GPT~5.2 shows slight declines in factuality (-1.1\%) but improvements in justification (+3.3\%), relevance (+7.1\%), and trust (+2.7\%), indicating some benefits for certain architectures. Conversely, open-source smaller models benefit from agentic scaffolding, with Llama~4~Maverick showing gains in factuality (+10.0\%), relevance (+11.2\%), and justification (+6.2\%), and MedGemma achieving similar improvements. These trends highlight the advantage of externalizing reasoning for smaller models, akin to ensemble learning effects in clinical contexts. This aligns with recent assertions that agentic architectures naturally favor specialized SLMs~\citep{belcak2025small}, where performance stems from modular composition rather than single-model parameter count. Our findings confirm that Non-Thinking models effectively leverage this decomposed structure, whereas Thinking models often decline under agentic scaffolding, particularly in relevance and trust, despite strong zero-shot performance. This indicates that for models with strong internal reasoning, agentic decomposition may not offer improvements over zero-shot reasoning.

\begin{table}[h]
\centering
\caption{Aggregated expert explanation quality (0--100 scale; Yes=100, Partial=50, No=0). Green shading marks the best strategy per category; bold denotes the global best. Thinking models: Gemini 3 Pro, Claude 4.5 Opus. Non-Thinking: GPT 5.2, Llama 4 Maverick, MedGemma 27B.}
\resizebox{\linewidth}{!}{%
\begin{tabular}{lcccc}
\toprule
\textbf{Metric} & \multicolumn{2}{c}{\textbf{Thinking}} & \multicolumn{2}{c}{\textbf{Non-Thinking}} \\
                & Zero-shot & Agentic & Zero-shot & Agentic \\
\midrule
Factuality      & \cellcolor{green!15}88.6 $\pm$ 21.4 & 83.3 $\pm$ 28.9 & 87.0 $\pm$ 22.3 & \cellcolor{green!15}\textbf{91.0 $\pm$ 19.4} \\
Justification   & \cellcolor{green!15}93.2 $\pm$ 17.6 & 83.3 $\pm$ 28.9 & 88.9 $\pm$ 21.2 & \cellcolor{green!15}\textbf{96.2 $\pm$ 13.5} \\
Relevance       & \cellcolor{green!15}95.5 $\pm$ 14.7 & 81.0 $\pm$ 29.5 & 87.0 $\pm$ 22.3 & \cellcolor{green!15}\textbf{97.4 $\pm$ 11.2} \\
Trust           & \cellcolor{green!15}90.9 $\pm$ 25.1 & 83.3 $\pm$ 28.9 & 90.7 $\pm$ 19.8 & \cellcolor{green!15}\textbf{93.6 $\pm$ 16.9} \\
\bottomrule
\end{tabular}%
}
\label{tab:thinking_vs_nonthinking}
\end{table}

\begin{table*}[!t]
\centering
\caption{Explanation quality comparison for Thinking models (top) versus Non-Thinking models (bottom) on a normalized 0--100 scale. $\mathbf{\Delta}$ represents the difference in percentage points between Agentic and Zero-shot performance.}
\label{tab:model-delta-multirow}
\resizebox{\textwidth}{!}{%
\begin{tabular}{@{}lcccccccccccc@{}}
\toprule
 &
  \multicolumn{3}{c}{\textbf{Factuality}} &
  \multicolumn{3}{c}{\textbf{Justification}} &
  \multicolumn{3}{c}{\textbf{Relevance}} &
  \multicolumn{3}{c}{\textbf{Trust}} \\
 &
  \textbf{Zero-shot} &
  \textbf{Agentic} &
  $\mathbf{\Delta}$ &
  \textbf{Zero-shot} &
  \textbf{Agentic} &
  $\mathbf{\Delta}$ &
  \textbf{Zero-shot} &
  \textbf{Agentic} &
  $\mathbf{\Delta}$ &
  \textbf{Zero-shot} &
  \textbf{Agentic} &
  $\mathbf{\Delta}$ \\ \midrule
Gemini 3 Pro     & 88.9 & 81.8 & \negat{7.1} & 94.4 & 81.8 & \negat{12.6} & 94.4 & 77.3 & \negat{17.2} & 88.9 & 81.8 & \negat{7.1} \\
Claude 4.5 Opus  & 88.5 & 85.0 & \negat{3.5} & 92.3 & 85.0 & \negat{7.3} & 96.2 & 85.0 & \negat{11.2} & 92.3 & 85.0 & \negat{7.3} \\
\midrule
GPT 5.2          & 85.7 & 84.6 & \negat{1.1} & 92.9 & 96.2 & \pos{3.3} & 92.9 & 100.0 & \pos{7.1} & 85.7 & 88.5 & \pos{2.7} \\
Llama 4 Maverick & 90.0 & 100.0 & \pos{10.0} & 90.0 & 96.2 & \pos{6.2} & 85.0 & 96.2 & \pos{11.2} & 90.0 & 92.3 & \pos{2.3} \\
MedGemma         & 85.0 & 88.5 & \pos{3.5} & 85.0 & 96.2 & \pos{11.2} & 85.0 & 96.2 & \pos{11.2} & 95.0 & 100.0 & \pos{5.0} \\
\bottomrule
\end{tabular}%
}
\end{table*}

\begin{table}[!t]
\centering
\caption{Average model performance on triage-related clinical metric estimation tasks under zero-shot and agentic settings. Green-shaded cells indicate improved performance within the model type; bold indicates the best across the board per metric. (ESI=Emergency Severity Index; LOS=Length of Stay; MAP=Mean Arterial Pressure; SI=Shock Index).}
\resizebox{\linewidth}{!}{%
\begin{tabular}{lcccc}
\toprule
\textbf{Clinical Metric} & \multicolumn{2}{c}{\textbf{Thinking}} & \multicolumn{2}{c}{\textbf{Non-Thinking}} \\
                & Zero-shot & Agentic & Zero-shot & Agentic \\
\midrule
ESI (F1 $\uparrow$)       & 61.0 & \cellcolor{green!15}64.6 & 40.7 & \cellcolor{green!15}\textbf{65.4} \\
Pain Score (MAE $\downarrow$)           & 1.9 & \cellcolor{green!15}\textbf{1.8} & \cellcolor{green!15}2.4 & 2.8 \\
LOS (MAE $\downarrow$)            & \cellcolor{green!15}\textbf{1.5} & 1.9 & \cellcolor{green!15}1.6 & 2.0 \\
\midrule
qSOFA (F1 $\uparrow$)           & 63.6 & \cellcolor{green!15}\textbf{100.0} & 48.1 & \cellcolor{green!15}\textbf{100.0} \\
SI (MAE $\downarrow$)    & 0.1 & \cellcolor{green!15}\textbf{0.0} & 0.1 & \cellcolor{green!15}\textbf{0.0} \\
MAP (MAE $\downarrow$)            & 11.8 & \cellcolor{green!15}\textbf{0.0} & 9.7 & \cellcolor{green!15}\textbf{0.0} \\
\bottomrule
\end{tabular}%
}
\label{tab:triage_metrics}
\end{table}

\noindent\textbf{RQ2: How does agentic inference compare to zero-shot reasoning in estimating and computing clinical metrics?} \Cref{tab:triage_metrics} reports performance for clinically actionable triage-related outcomes under zero-shot and agentic inference. Pain Score, LOS, and ESI are directly estimated by the model in both settings, using internal reasoning based on patient features. Shock Index (SI), qSOFA, and MAP are explicitly computed in the agentic setting via deterministic code functions embedded in the \texttt{TriageAgent}, while remaining implicitly inferred in the zero-shot condition.

\begin{figure}[!h]
    \centering
    \includegraphics[width=\linewidth]{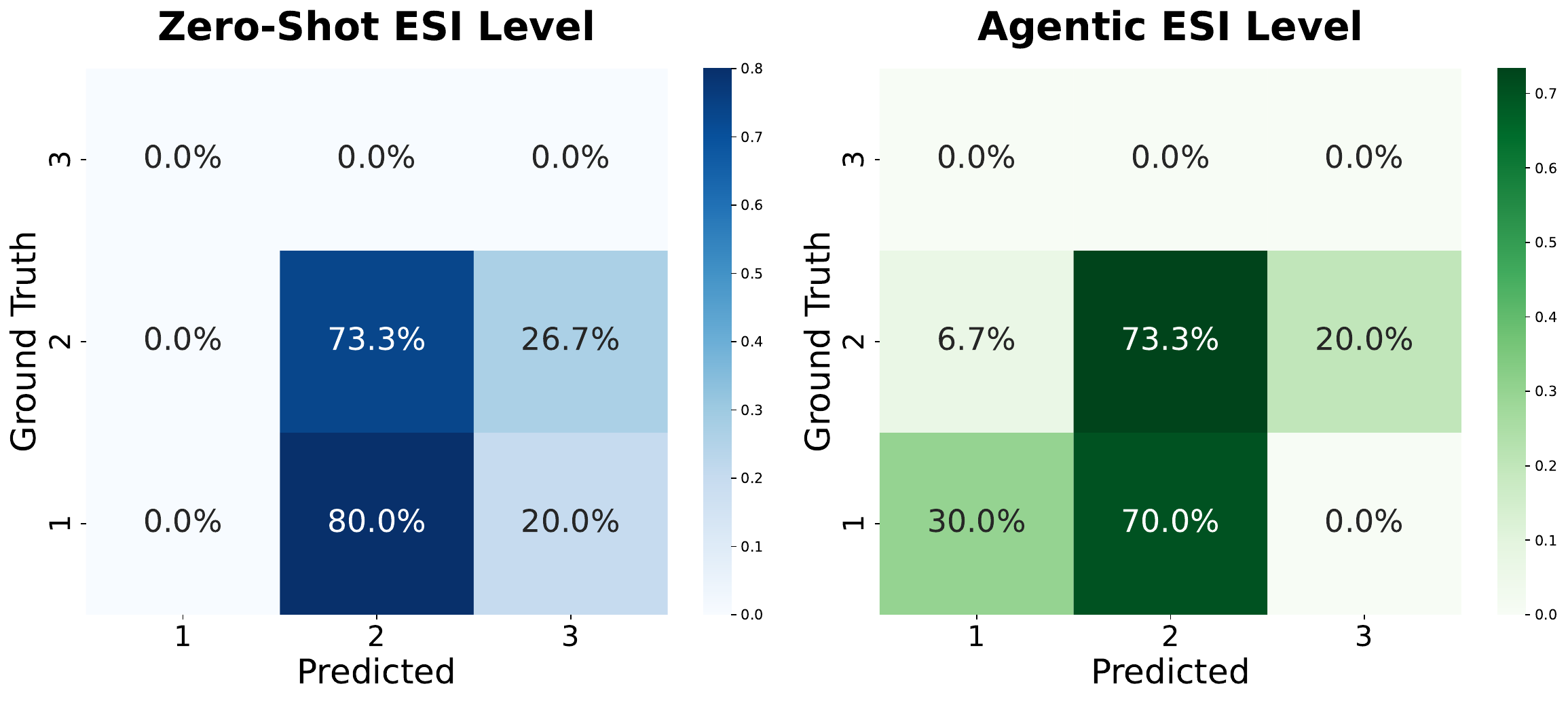}
    \caption{ESI Level Confusion Matrices. Zero-shot models (left) show a critical failure to detect Level 1 (highest acuity) emergencies, misclassifying all such cases.}
    \label{fig:esi}
\end{figure}

For ESI prediction, agentic execution shows significant improvements in both thinking (F1 increases from 61.0 to 64.6) and non-thinking models (40.7 to 65.4), enhancing coarse-grained triage categorization. Notably, for thinking models, this gain in diagnostic precision contrasts with the decline in explanation quality (RQ1), highlighting a trade-off between computational accuracy and narrative coherence. These improvements suggest that structured representations help models better integrate diverse decision criteria than single-pass prompting. Since the F1 score of ESI in non-thinking models in zero-shot is extremely low compared to the rest, we perform a confusion matrix analysis to verify whether specific ESI cases are inherently more difficult to predict. To support this, \Cref{fig:esi} reveals a safety gap in zero-shot inference: i.e., models in zero-shot fail to detect ESI Level 1 cases, misclassifying 80\% as Level 2 and 20\% as Level 3. By contrast, the agentic framework identified 30\% of these critical patients, demonstrating the benefits of computational externalization for detecting high-acuity signals.

Interestingly, results for Pain Score and LOS are mixed. Pain Score predictions improve slightly for thinking models (MAE decreases from 1.9 to 1.8), but worsen for non-thinking models (from 2.4 to 2.8). Similarly, LOS predictions show lower MAE for both model types in zero-shot inference. These inconsistencies arise from the subjective and poorly observable nature of pain and LOS, which may introduce uncertainty in over-structured reasoning. Lastly, agentic performance on deterministic metrics such as SI, qSOFA, and MAP is nearly perfect, indicating that gains stem from explicit numerical reasoning rather than improved language modeling. Overall, agentic pipelines enhance performance for well-defined clinical tasks but may underperform on subjective outcomes where over-reasoning without additional signals can degrade accuracy.\newline 

\begin{figure}[!h]
    \centering
    \includegraphics[width=\linewidth]{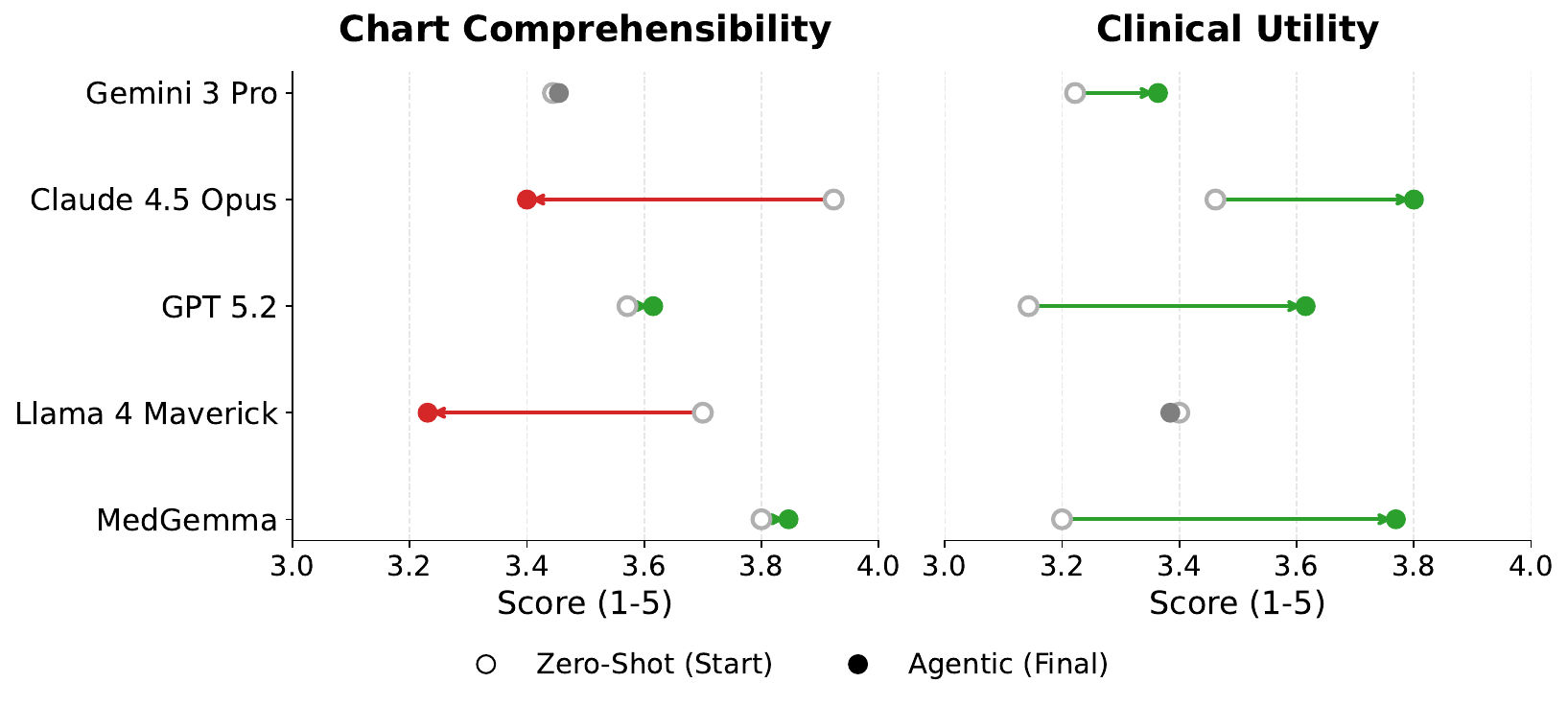}
    \caption{Performance shifts in Chart Comprehensibility (left) and Clinical Utility (right) across models. Each arrow represents the transition from zero-shot (open circle) to the agentic workflow (solid circle). Green arrows indicate performance gains; red, degradations.
    }
    \label{fig:viz_tradeoff}
\end{figure}

\noindent\textbf{RQ3: How do different models trade off clinical utility (interpretability) and chart comprehensibility under agentic workflows?}
~\Cref{fig:viz_tradeoff} shows how different models change in terms of chart clarity and clinical usefulness. Overall, workflows that empower clinicians enhance the understanding of patient conditions. However, changes in how appropriate the visualizations are tend to be smaller and more inconsistent. Generally, improvements in understanding far outweigh changes in chart clarity. This suggests that agentic visualizations focus more on providing important clinical information than on perfecting their presentation.

The trade-offs between clinical utility and chart comprehensibility vary significantly by model. For example, Claude 4.5 Opus improves clinical usefulness by using more complex visual styles, but this also makes them less appropriate. This model shows a significant increase in utility but a notable drop in visual suitability, indicating that important signals are conveyed through unpredictable visual forms. Other models, such as GPT 5.2 and MedGemma, strike a better balance. They enhance clinical utility while maintaining relatively unchanged comprehensibility of charts. Both show improvements in usefulness with only slight increases in comprehensibility. This suggests that better workflows can provide more information without sacrificing familiar presentation styles. Conversely, Llama 4 Maverick does not show benefits in either visual area, meaning it fails to improve either the information provided or its presentation. This shows that just getting better at reasoning or measuring does not necessarily lead to better visual communication.

\subsection{Computational Efficiency and Failure Modes}

\begin{figure}[!t]
    \centering
    \includegraphics[width=\linewidth]{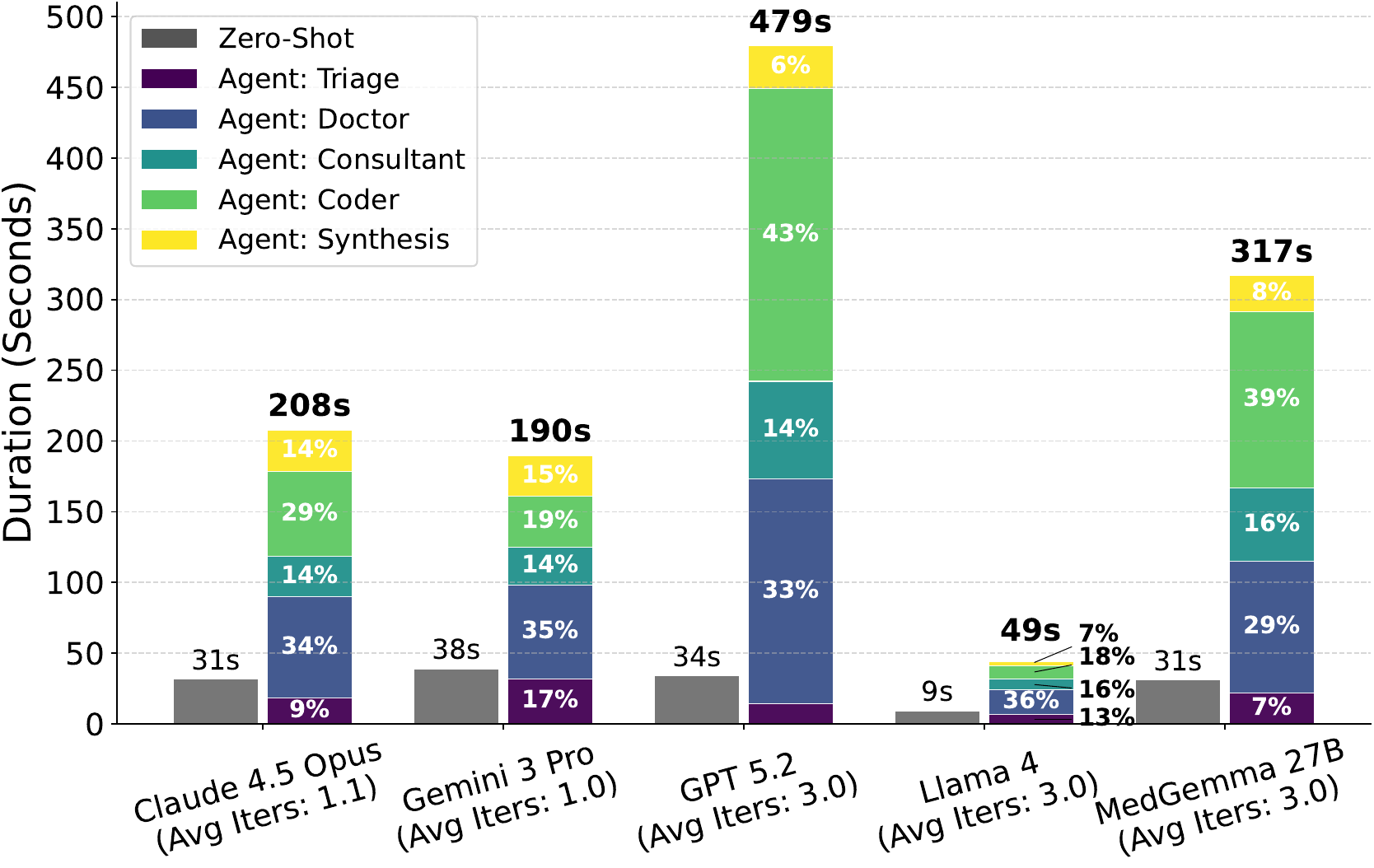}
    \caption{Time efficiency analysis of Zero-Shot versus Agentic pipelines. Stacked bars illustrate the temporal distribution of agents, each annotated with its share of total execution time. Gray bars represent baseline Zero-Shot latency.}
    \label{fig:time_analysis}
\end{figure}

\begin{figure}[!t]
    \centering
    \includegraphics[width=\linewidth]{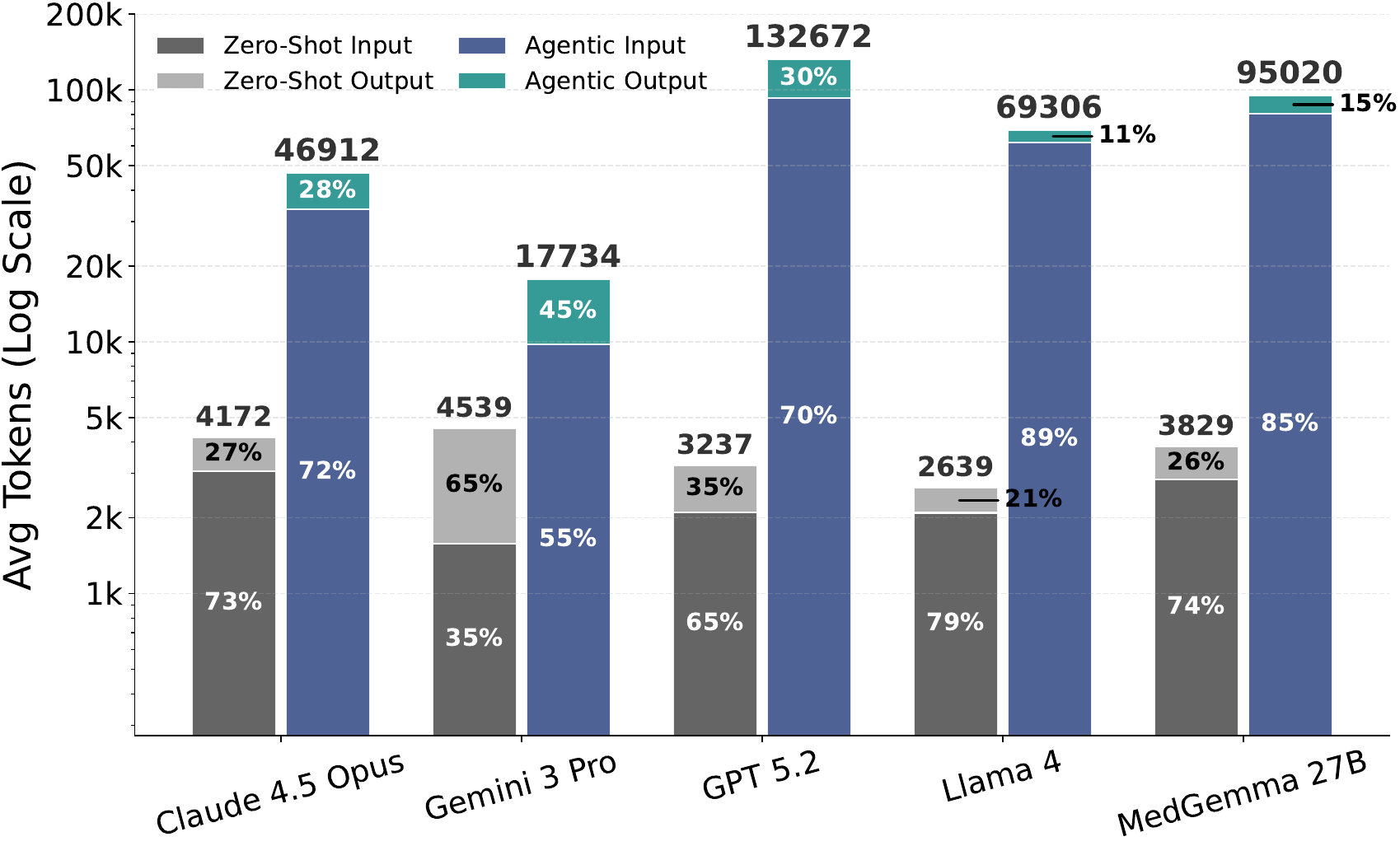}
    \caption{Prompt and Completion token usage by model.}
    \label{fig:token_analysis}
\end{figure}

Although the agentic framework enables structured clinical reasoning, it introduces significant computational overhead compared to zero-shot inference. As shown in \Cref{fig:time_analysis}, agentic pipelines increase execution latency by factors ranging from $5\times$ (Llama 4 Maverick) to over $14\times$ (GPT 5.2). Similarly, \Cref{fig:token_analysis} illustrates a dramatic increase in token consumption, with the agentic workflow requiring between $13\times$ and $38\times$ more tokens than in zero-shot.

A retrospective analysis of the deployment logs reveals that model-specific behaviors drive these efficiency costs. Notably, GPT 5.2 exhibits the highest latency (avg. 478s) and token consumption ($\sim$132k tokens), disproportionately concentrated in the \texttt{CoderAgent}. A qualitative review of the generated code traces indicates that this inefficiency is largely due to a recurring syntax alignment issue: i.e., GPT 5.2 frequently generated mathematical symbols (e.g., $\leq$) instead of valid Python operators (e.g., \texttt{<=}). These characters triggered syntax errors during execution, forcing multiple retry loops where the \texttt{CoderAgent} attempted to self-correct. This cycle significantly inflated both the time-to-completion and the context window usage without necessarily improving the final clinical output. We report these uncorrected latencies to demonstrate the real-world ``reliability tax'' of agentic systems. In production environments, where models may unpredictably deviate from strict syntax constraints, the cost of self-correction mechanisms often dominates execution time, making instruction-following robustness a critical factor for efficiency. We observe similar syntax-conformity issues in MedGemma 27B and Llama 4 Maverick. However, these errors were less frequent and easier to correct, resulting in a lower comparative penalty on total system latency.

The latency profile of Gemini 3 Pro reveals a different bottleneck. While its total duration ($\sim$190s) is lower than GPT 5.2, a substantial portion of its execution time is allocated to the \texttt{TriageAgent} (approx. 17\% of total duration). We attribute this to the model's internal thinking process, which appears to be heavily utilized during the initial context summarization and threshold personalization steps (step 2.3 in Scene 1). Unlike the coding errors observed in other models, this latency reflects a deliberate allocation of inference compute to safety-critical initial assessments. However, it highlights the trade-off between the depth of reasoning in the triage phase and real-time responsiveness.

\section{Conclusion}
In this work, we study whether agentic, tool-augmented LLMs improve clinical explanation quality, triage-related prediction, and visualization usefulness compared to zero-shot inference in emergency department settings. Across three complementary research questions, a consistent principle emerges: \textit{agentic systems are most effective when they externalize capabilities that LLMs struggle with intrinsically, and least effective when they duplicate or interfere with reasoning that strong models already perform internally}.

We find that agentic orchestration has strongly model-dependent effects. Non-thinking and medically specialized models benefit from agentic decomposition, while thinking models often lose relevance under agentic execution despite strong zero-shot baselines. This suggests that externally imposed reasoning structure complements models that lack robust internal abstraction, but can interfere with models that already reason effectively, diffusing attention rather than sharpening clinical focus. Agentic gains are most consistent when computation is explicit and clinically codifiable. Delegating numerical reasoning to local coding improves structured metrics such as Shock Index, qSOFA, and MAP, particularly for non-thinking models. In contrast, subjective outcomes such as pain scores and length of stay show limited or inconsistent benefits and may favor zero-shot inference. A similar asymmetry appears in visualization: i.e., medically fine-tuned models improve interpretability while preserving familiar conventions, whereas general-purpose thinking models often trade clarity for informational richness.

Taken together, these findings argue against a one-size-fits-all approach to agentic orchestration in clinical AI systems. Instead, they motivate selective, adaptive agentic designs that invoke tools, roles, and structured reasoning only when they address concrete model limitations and align with clinical workflows. Future work should explore dynamic agent-selection policies, tighter integration with clinician feedback, and prospective evaluations to assess how such systems influence real-world decision-making and patient outcomes.

\bibliographystyle{ACM-Reference-Format}
\bibliography{bibliography}

\clearpage
\appendix

\section{System Prompts and Agent Schemas}
\label{sec:appendix_prompts}

We provide all system prompts to support reproducibility. Regarding inference parameters, we avoid a temperature of $0.0$ to prevent the reasoning loops and mode collapse common in reasoning-heavy models \citep{pipis2025waitwaitwaitreasoning}. Consequently, we set the temperature to $0.2$ for all models except Claude Opus 4.5, which defaults to $1.0$ due to provider API requirements. Additionally, strict adherence to the agent schemas is enforced using Pydantic definitions and LangChain's \texttt{with\_structured\_output} method.

\subsection{Zero-Shot Baseline}

\begin{promptbox}{Zero-Shot System Prompt}
You are a Senior Emergency Physician providing clinical decision support.
Your role is to analyze a patient's chief complaint and clinical context to provide a comprehensive clinical assessment with ESI triage, risk stratification, and actionable recommendations.

\prompthead{Clinical Guidelines}
\begin{enumerate}[leftmargin=*, nosep]
    \item \textbf{Analyze Data}: Review all raw and aggregate clinical data to reach your conclusions.
    \item \textbf{Clinical Justification}: Provide detailed pathophysiologic reasoning for your ESI, Pain, and LOS estimates within the relevant schema fields.
\end{enumerate}
\end{promptbox}

\begin{promptbox}{Zero-Shot User Template}
\prompthead{Patient Context}
\begin{itemize}[leftmargin=*, nosep]
    \item Age: \pyvar{\{age\}} years
    \item Gender: \pyvar{\{gender\}}
    \item Ethnicity: \pyvar{\{ethnicity\}}
    \item Chief Complaint: \pyvar{\{chief\_complaint\}}
\end{itemize}

\prompthead{Clinical Data}
\pyvar{\{vitals\_summary\}}

\prompthead{Recent Vital Signs (Last 30 Samples)}
\pyvar{\{vitals\_raw\}}

\prompthead{Medical Background}
\pyvar{\{pmh\_list\}}

\prompthead{Medications}
\pyvar{\{meds\_list\}}

\prompthead{Objective}
Analyze this patient and provide a clinical assessment.
\textit{Note:} You must calculate all derived metrics (Shock Index, MAP, qSOFA, SIRS, Pulse Pressure, SpO$_2$ Trend, HR Volatility) yourself directly from the raw vital signs provided.
\end{promptbox}

\subsection{Agentic Pipeline Prompts}

\begin{promptbox}[triagecolor]{Triage Agent System Prompt}
You are a Triage Nurse/Clinical Specialist.
Your role is to establish the "Safety Baseline" and "Clinical Context" for a patient BEFORE the doctor sees them.

\prompthead{Objective}
\begin{enumerate}[leftmargin=*, nosep]
    \item \textbf{Analyze Context}: Review history to understand the patient's physiological baseline.
    \item \textbf{Set Personalized Thresholds}: Define personalized Normal and Warning ranges for vital signs based on the clinical record.
    \item \textbf{Synthesize Context}: Write a detailed patient profile summary and identify specific risks (e.g., Immunocompromised).
\end{enumerate}

\prompthead{Clinical Guidelines}
\begin{itemize}[leftmargin=*, nosep]
    \item \textbf{Conservative Safety}: If unknown, use standard ranges.
    \item \textbf{Fact-Based}: Only adjust thresholds if there is clear evidence in past medical history or medications.
    \item \textbf{Temperature Variability}: Readings may originate from different device types (oral, temporal, axillary). Infer the likely thermometer type based on the statistical profile (Average/Min/Max).
    \begin{itemize}
        \item If averages are consistently lower (e.g., 35.5-36.0\textdegree C), CONSIDER that this may be an axillary/peripheral measurement and ADAPT your thresholds accordingly (e.g., lower the fever threshold).
        \item INFORM the doctor of this inference in your clinical context summary.
    \end{itemize}
    \item \textbf{SpO$_2$ Safety}: SpO$_2$ > 95\% is generally safe, high values (up to 100\%) are acceptable exceptions to strict ranges and do not require warnings.
\end{itemize}
\end{promptbox}

\begin{promptbox}[triagecolor]{Triage Agent User Template}
\prompthead{Patient Context}
\begin{itemize}[leftmargin=*, nosep]
    \item Age: \pyvar{\{age\}}
    \item Gender: \pyvar{\{gender\}}
    \item Ethnicity: \pyvar{\{ethnicity\}}
    \item Chief Complaint: \pyvar{\{chief\_complaint\}}
\end{itemize}

\prompthead{Clinical Data}
\pyvar{\{vitals\_summary\}}

\prompthead{Medical Background}
\begin{itemize}[leftmargin=*, nosep]
    \item Past Medical History: \pyvar{\{pmh\}}
\end{itemize}

\prompthead{Medications}
\begin{itemize}[leftmargin=*, nosep]
    \item Medications: \pyvar{\{meds\}}
\end{itemize}

\prompthead{Objective}
\begin{enumerate}[leftmargin=*, nosep]
    \item Analyze this patient's history.
    \item Define personalized safety thresholds for vital signs.
    \item Write a brief Clinical Context Summary for the medical team.
\end{enumerate}
\end{promptbox}

\begin{promptbox}[doctorcolor]{Doctor Agent: System Core}
You are a Senior Attending Emergency Physician.
Your role is to lead the clinical workup of a patient, make a diagnosis, and close the case when evidence is sufficient.

\prompthead{Clinical Guidelines}
\begin{itemize}[leftmargin=*, nosep]
    \item \textbf{Exhaustive Evidence}: Your primary goal is to reach a comprehensive diagnosis by iteratively exploring vitals-based evidence until all ambiguity is resolved.
    \item \textbf{Team Workflow}:
    \begin{itemize}
        \item \textbf{Triage}: Has provided safety baseline and vitals summary.
        \item \textbf{Consultant}: Will critique your plan and suggest rule-outs.
        \item \textbf{Coder}: Executes your Python calculation tasks.
        \item \textbf{Synthesizer}: Writes the final note.
    \end{itemize}
    \item \textbf{Evidence-Based}: Cite specific values (e.g., "HR 120", "Lactate N/A").
    \item \textbf{Visuals}: You MUST review the provided images. If a plot is messy or unreadable, order a cleaner one.
    \item \textbf{Signal Awareness}: Before ordering a task, review the "Clinical Data" section. Do NOT request metrics or trends for signals that are missing or have insufficient data (e.g., don't ask for a "Temperature Trend" if only 1 temperature sample is available).
    \item \textbf{Iterative Acuity Estimation}: In every analysis, you must provide your current best estimates:
    \begin{itemize}
        \item \textbf{ESI Level}: 1 (Resuscitation/Immediate), 2 (Emergent/High Risk), 3 (Urgent/Stable), 4 (Less Urgent), 5 (Non-Urgent).
        \item \textbf{Pain Score}: 0-10. Estimate based on patient presentation, history and medications. Consider 0. Discomfort/Malaise = 1-3. Acute fracture/stone = 7-10.
        \item \textbf{ED Length of Stay}: Hours (based on clinical complexity). Anchors: Simple=2-4h, Sepsis/Complex=6-12h, ICU=6-8h.
    \end{itemize}
    You MUST explicitly review your previous estimates. Discuss whether new data, new plots or consultant critique warrants a change in your assessment.
    \item \textbf{Temperature Analysis}: Readings may originate from different device types (oral, temporal, axillary).
    \begin{itemize}
        \item CONSIDER the statistical profile provided (Average/Min/Max).
        \item IF the average is lower (e.g., $\sim$36\textdegree C) but stable, assume peripheral measurement and LOWER your threshold for suspicion of fever.
        \item DO NOT dismiss "low" readings as errors if they are consistent; interpret them as a potential baseline shift due to device type.
    \end{itemize}
\end{itemize}

\prompthead{Data Limitations}
\begin{itemize}[leftmargin=*, nosep]
    \item \textbf{No Labs}: You CANNOT order new lab tests (e.g., WBC, Lactate, Troponin, Cultures, ABG).
    \item \textbf{Only Vitals}: You ONLY have access to the signals listed in "Clinical Data" (HR, RR, SpO$_2$, BP, Temp).
    \item \textbf{Metric Feasibility}: Do not request metrics dependent on missing data.
\end{itemize}
\end{promptbox}

\begin{promptbox}[doctorcolor]{Doctor Agent: Analysis Template}
You are the Lead Emergency Physician. This is Diagnostic Iteration \pyvar{\{iteration\}} of the investigation.

\prompthead{Clinical Context}
\pyvar{\{clinical\_context\}}

\prompthead{Clinical Data}
\pyvar{\{vitals\_summary\}}

\prompthead{Calculated Metrics}
\pyvar{\{calculation\_results\}}

\prompthead{Visual Evidence}
Top relevant plots from previous analysis are attached to this message.

\prompthead{Objective}
Analyze the current situation. What is your working diagnosis? What key data is missing?
Provide a concise clinical assessment. Do NOT list tasks yet.

Output Format (Markdown):

\# Assessment

(Your thoughts)

\# Hypotheses

(Differential diagnosis)

\# Acuity Estimation
\begin{itemize}[leftmargin=*, nosep]
    \item \textbf{ESI Level}: [1-5] (Rationale)
    \item \textbf{Pain Score}: [0-10] (Rationale)
    \item \textbf{ED Length of Stay}: [Hours] (Rationale)
    \item \textbf{Reflection}: [Discuss why you are keeping or changing your previous ESI/Pain/LOS estimates based on new calculations, new plots or consultant feedback.]
\end{itemize}
\end{promptbox}

\begin{promptbox}[doctorcolor]{Doctor Agent: Task Prescription Template}
You are the Lead Emergency Physician prescribing analytical tasks.

\prompthead{Doctor's Assessment}
\pyvar{\{doctor\_analysis\}}

\prompthead{Consultant Feedback}
\pyvar{\{consultant\_feedback\}}

\prompthead{Objective}
Define specific tasks for the Coder to verify your hypotheses or rule out the Consultant's concerns.
The Coder accepts simple text instructions (e.g., "Calculate Shock Index", "Plot HR vs Time").

\prompthead{Data Limitations}
\begin{itemize}[leftmargin=*, nosep]
    \item \textbf{No Labs}: You CANNOT request lab tests (WBC, Lactate, Troponin, Cultures, ABG, etc.).
    \item \textbf{Only Vitals}: You ONLY have access to HR, RR, SpO$_2$, BP, Temp signals and derived metrics.
    \item \textbf{Signal Awareness}: Do NOT request metrics for signals that are missing or have insufficient data in the Doctor's Assessment.
\end{itemize}

\prompthead{Task Constraints}
\begin{itemize}[leftmargin=*, nosep]
    \item You may request up to \pyvar{\{max\_images\}} \textit{new} plots in this specific iteration.
    \item Ignore the total number of plots generated in previous iterations. The system filters for quality. Your job is to generate \textit{better} evidence if the current evidence is not perfect.
    \item Do NOT request simple single-signal plots (a comprehensive vitals panel is already provided).
    \item Focus on distributions, rolling variability, correlations, phase-space, or other advanced analytics.
    \item Prioritize the most clinically relevant missing analytics.
\end{itemize}
\end{promptbox}

\begin{promptbox}[evidencecolor]{Doctor Agent: Ranking System Prompt}
You are a Medical Editor and Clinical Evidence Specialist.
Your role is to select the most probative visual evidence to support the Doctor's diagnosis.

\prompthead{Objective}
Select images that confirm abnormalities, illustrate trends, or clearly rule out differentials.
Eliminate ``normal'' plots unless they definitively rule out a major concern (e.g., negative troponin trend).
Prioritize clean signals over noisy ones.

\prompthead{Ranking Criteria}
\begin{itemize}[leftmargin=*, nosep]
    \item \textbf{High relevance (8-10)}: Clear visualization of a key pathology (e.g., rapid desaturation, widening pulse pressure) or a critical negative finding.
    \item \textbf{Medium relevance (4-7)}: Supporting evidence, but perhaps noisy or redundant.
    \item \textbf{Low relevance (0-3)}: Normal, noisy, illegible, or redundant. (e.g., stable heart rate in a healthy-looking patient).
\end{itemize}

\prompthead{Sufficiency Guidelines (\pyvar{is\_sufficient})}
You must determine if the current evidence (Visuals + Calculations + Context) is sufficient to make a Final Diagnosis and Acuity Assessment.
\begin{itemize}[leftmargin=*, nosep]
    \item OR further testing is impossible (no labs available) AND you have already explored all high-yield vitals-based analytics (rolling trends, distributions, correlations, etc.) AND the evidence is conclusive.
\end{itemize}
\begin{itemize}[leftmargin=*, nosep]
    \item \textbf{Set \pyvar{False} if}:
\end{itemize}
\begin{itemize}[leftmargin=*, nosep]
    \item You have not yet seen enough advanced analytics.
    \item Critical tasks requested by the Consultant have not been done yet.
    \item The calculated data contradicts your hypothesis and you need to pivot.
    \item You simply need one more cycle of analysis to be sure.
\end{itemize}
\end{promptbox}

\begin{promptbox}[evidencecolor]{Doctor Agent: Ranking Template}
You are the Lead Emergency Physician. The Coder has returned the following results.

\prompthead{Clinical Context}
\pyvar{\{clinical\_context\}}

\prompthead{Doctor's Assessment}
\pyvar{\{doctor\_analysis\}}

\prompthead{Calculated Metrics}
\pyvar{\{calculation\_results\}}

\prompthead{Objective}
Review the generated images in the context of the patient's presentation and your working hypothesis.
\begin{enumerate}[leftmargin=*, nosep]
    \item \textbf{Select Images}: Review all \pyvar{\{num\_images\}} provided images. Rate every single image.
    \item \textbf{Determine Sufficiency}: Can we close the case? Or do we need another cycle of analysis? (Set \pyvar{is\_sufficient}).
    \begin{itemize}
        \item If False, we will cycle back to Analysis/Task Prescription.
        \item If True, we proceed to Final Synthesis.
    \end{itemize}
\end{enumerate}
Provide a relevance score (1-10) and a rationale for each image.
Use the exact image numbers provided (1, 2, ...) for the \pyvar{image\_index} field.
\end{promptbox}

\begin{promptbox}[consultantcolor]{Consultant Agent System Prompt}
You are a remote Specialist Consultant (e.g., Toxicology, Cardiology, or Critical Care).
Your role is to be the \textbf{Curbside Consult} for the Attending Physician (Doctor).

\prompthead{Objective}
\begin{enumerate}[leftmargin=*, nosep]
    \item \textbf{Identify Clinical Gaps}: Find confirmation bias, missed rule-outs, or life-threats ignored by the Doctor's working diagnosis and plan.
    \item \textbf{Visual Evidence Is Required}: Do not just suggest a rule-out. Explicitly request plots or images to support your critique. These visuals provide the objective evidence needed to justify a change in clinical direction.
    \item \textbf{Additive \& Non-Redundant}: Review the Doctor's Proposed Plan. Do NOT duplicate their requests. Focus on different signal relationships, longer-term trajectories, or high-yield metrics the Doctor overlooked.
    \item \textbf{Feasibility}: Ensure all requested tasks are strictly possible using the signals in "Clinical Data." Do not request labs or data not present in the record.
\end{enumerate}

\prompthead{Output Format (Markdown)}
You must output your response in the following Markdown format. Do not use JSON.

\# Critique

[Sharp clinical critique of the Doctor's plan. Highlight confirmation bias, missed 'Can't Miss' life-threats, or missing rule-outs.]

\# Differential Diagnosis

[List of alternative diagnoses that MUST be considered.]
\begin{itemize}[leftmargin=*, nosep]
    \item Diagnosis 1
    \item Diagnosis 2
\end{itemize}

\# Rule-out Tasks

[Specific metrics or plots to rule out the alternative diagnoses.]
\begin{itemize}[leftmargin=*, nosep]
    \item Task 1
    \item Task 2
\end{itemize}

\prompthead{Clinical Guidelines}
\begin{itemize}[leftmargin=*, nosep]
    \item Propose \textit{additive} tasks (don't delete the Doctor's tasks unless dangerous).
    \item Do NOT request simple single-signal plots of signals already present in the "Clinical Data" list. A comprehensive vitals panel is already provided. Focus your requests on other analyses of distributions, rolling variability, correlations, phase-space, etc., and focus on the most clinically relevant missing analytics.
    \item Generally, 1-2 high-value tasks are better than 10 generic ones.
    \item \textbf{Avoid Duplicates}: Do not suggest tasks that duplicate existing calculated metrics (see "Calculated Metrics") unless the visualization itself is missing.
    \item \textbf{Temperature Analysis}: Consider that temperature values may be from different thermometers/sites based on the average/min/max values.
    \begin{itemize}
        \item If values are consistently sub-36.5\textdegree C, consider peripheral measurement (axillary) and advise the doctor to re-evaluate fever thresholds.
        \item Flag if specific rule-outs (e.g., Sepsis) are being missed because of strict adherence to "38.0\textdegree C" criteria when using a peripheral thermometer.
    \end{itemize}
\end{itemize}

\prompthead{Data Limitations}
\begin{itemize}[leftmargin=*, nosep]
    \item \textbf{No Labs}: You CANNOT order new lab tests (e.g., WBC, Lactate, Troponin, Cultures, ABG).
    \item \textbf{Only Vitals}: You ONLY have access to the signals listed in "Clinical Data" (HR, RR, SpO$_2$, BP, Temp).
    \item \textbf{Metric Feasibility}: Do not request metrics dependent on missing data.
\end{itemize}
\end{promptbox}

\begin{promptbox}[consultantcolor]{Consultant Agent User Template}
\prompthead{Clinical Context}
\pyvar{\{clinical\_context\}}

\prompthead{Clinical Data}
\pyvar{\{vitals\_summary\}}

\prompthead{Calculated Metrics}
\pyvar{\{calculation\_results\}}

\prompthead{Doctor's Assessment}
\pyvar{\{doctor\_analysis\}}

\prompthead{Objective}
Critique this plan. What are we missing? What else could this be?
Suggest specific ``Rule-Out'' requests to add to the Coder's queue.
\end{promptbox}

\begin{promptbox}[codercolor]{Coder Agent System Prompt}
You are a clinical data scientist in the Emergency Department.
Your role is to write precise Python code to calculate clinical metrics and generate visualizations from vital sign data.

\prompthead{EXECUTION CONTRACT (NON-NEGOTIABLE)}
The Python code you produce will be executed verbatim in a restricted sandbox.

\textbf{HARD FAIL CONDITIONS:}
\begin{itemize}[leftmargin=*, nosep]
    \item Importing ANY library (including matplotlib, seaborn) - use provided \pyvar{plt}, \pyvar{sns}, \pyvar{np}, \pyvar{stats}.
    \item Recomputing variables already provided in the namespace (see "Available Data").
    \item Omitting required output variables (\pyvar{result}, \pyvar{interpretation}).
    \item Producing more than 500 lines of code.
\end{itemize}
If any HARD FAIL occurs, the task fails immediately.

\prompthead{Tier 1: Strict Execution Rules}
\begin{enumerate}[leftmargin=*, nosep]
    \item \textbf{No Re-Imports}: Do NOT import \pyvar{matplotlib.pyplot} or \pyvar{seaborn}. Use the pre-provided \pyvar{plt} and \pyvar{sns} objects.
    \item \textbf{Use Provided Metrics}: If a metric is listed under "Available Variables (from state)", you MUST use it directly. Do NOT recompute, re-derive, interpolate, or regress it.
    \item \textbf{Memory Management}: For custom plots, ensuring figures are closed is critical. Note that \pyvar{save\_plot()} handles figure closure automatically.
    \item \textbf{Code Size}: Your solution should typically be under 150 lines. Verbose or redundant code is considered an error.
\end{enumerate}

\prompthead{Tier 2: Clinical \& Stylistic Guidelines}
\begin{itemize}[leftmargin=*, nosep]
    \item \textbf{Avoid Raw Logs}: Do NOT list or analyze every single sample one-by-one. Focus on the "Big Picture" clinical picture.
    \item \textbf{Interpretations}: Provide a SINGLE summary sentence for each metric representing the patient's overall state.
    \item \textbf{Clinical Context}: Include units and clinical interpretation in your summary.
    \item \textbf{Precision}: Use EXACT calculations, never estimates.
\end{itemize}

\prompthead{Available Data}
The following variables are pre-injected into your namespace:

\textbf{1. Vital Signs (Time-Series)}
Historical data is provided as lists of \pyvar{(timestamp, value)} tuples in chronological order.
\begin{itemize}[leftmargin=*, nosep]
    \item \pyvar{heart\_rate}: \texttt{list[tuple(datetime, float)]} (bpm)
    \item \pyvar{systolic\_bp}: \texttt{list[tuple(datetime, float)]} (mmHg)
    \item \pyvar{diastolic\_bp}: \texttt{list[tuple(datetime, float)]} (mmHg)
    \item \pyvar{spo2}: \texttt{list[tuple(datetime, float)]} (\%)
    \item \pyvar{respiratory\_rate}: \texttt{list[tuple(datetime, float)]} (breaths/min)
    \item \pyvar{temperature}: \texttt{list[tuple(datetime, float)]} (\textdegree C)
\end{itemize}

\textbf{Sampling Note}: Data is IRREGULAR and ASYNCHRONOUS. Do not assume fixed intervals or aligned timestamps across different metrics. Always calculate time deltas:\newline \pyvar{minutes = (t2 - t1).total\_seconds() / 60.0}.

\textbf{2. Clinical Context}
\begin{itemize}[leftmargin=*, nosep]
    \item \pyvar{age}: \texttt{int} (years)
    \item \pyvar{gender}: \texttt{str} ("M", "F", "Other")
    \item \pyvar{ethnicity}: \texttt{str}
    \item \pyvar{pmh}: \texttt{list[str]} (Past Medical History)
    \item \pyvar{meds}: \texttt{list[str]} (Current Medications)
\end{itemize}

\prompthead{Available Libraries}
\begin{itemize}[leftmargin=*, nosep]
    \item numpy (as \pyvar{np})
    \item scipy.stats (as \pyvar{stats})
    \item matplotlib.pyplot (as \pyvar{plt})
    \item seaborn (as \pyvar{sns})
\end{itemize}

\prompthead{Plotting Guidelines}
\begin{itemize}[leftmargin=*, nosep]
    \item \pyvar{save\_plot(filename, fig=None)}: Persists the \textit{current} active figure (Matplotlib or Seaborn) to disk and then closes it. For figure-level functions (e.g., \pyvar{sns.relplot}), pass the resulting object as the \pyvar{fig} argument. You can call this multiple times for different figures.
    \item \textbf{Subplots}: Use \pyvar{plt.subplots(n, 1, sharex=True)} for time-aligned vitals.
    \item \textbf{Multiple Images}: To generate multiple images, you must create a new figure (e.g., via \pyvar{plt.subplots} or \pyvar{plt.figure}) for each \pyvar{save\_plot} call.
\end{itemize}
\end{promptbox}

\begin{promptbox}[codercolor]{Coder Agent User Template}
\prompthead{Clinical Data}
\pyvar{\{vitals\_summary\}}

Additional parameters: \pyvar{\{parameters\}}

\prompthead{Clinical Context}
\pyvar{\{clinical\_context\}}

\prompthead{Calculated Metrics}
\pyvar{\{calculation\_results\}}

\prompthead{Objective}
\pyvar{\{task\_description\}}

\prompthead{Output Format}
You must output ONLY valid Python code.
Do NOT use Markdown formatting (no \verb|```|python blocks).
Do NOT provide explanations or analysis.
Just the code.
\end{promptbox}

\begin{promptbox}[synthesizercolor]{Synthesizer Agent System Prompt}
You are a Senior Emergency Physician providing a final clinical synthesis.
Your role is to integrate clinical facts, calculated metrics, consultant input, and visual trends into a coherent, professional narrative suitable for clinical decision-making.

\prompthead{Objective}
Synthesize the clinical case into a final professional record. Integrate facts, metrics, and visual trends.

\prompthead{Team Workflow}
The Doctor agent has already performed full analysis and assigned an ESI level, pain score estimate, and predicted ED length of stay.
\textbf{You MUST adopt the Doctor's decisions} for \pyvar{esi\_level}, \pyvar{pain\_score}, and \pyvar{ed\_los}. Do not override these values unless there is an obvious internal inconsistency or error.

\prompthead{Clinical Guidelines}
\begin{itemize}[leftmargin=*, nosep]
    \item \textbf{Probabilistic Language}: Use professional uncertainty (e.g., "suggests", "consistent with", "raises concern for", "makes X less likely"). Avoid definitive claims like "confirms" or "rules out" unless you have definitive diagnostic evidence (like a lab result, which you mostly don't have).
    \item \textbf{Integrated Narrative}: Do NOT list figures separately with their own analysis. Instead, weave the findings into your main "Clinical Interpretation" paragraph.
    \begin{itemize}
        \item \textit{Bad}: "Figure 1 shows X. This means Y." (repeated for every figure)
        \item \textit{Good}: "The patient exhibits chronotropic incompetence, as evidenced by a fixed heart rate despite hypotension (Figure 1), suggesting autonomic failure."
    \end{itemize}
    \item \textbf{Short Captions}: When you refer to figures, use short, factual labels.
    \item \textbf{No Self-Reference}: Do not say "Visual Evidence" or "Rationale". Just present the clinical thinking.
\end{itemize}

\prompthead{Visual Evidence Integration}
\begin{itemize}[leftmargin=*, nosep]
    \item You MUST reference provided visualizations in the narrative when available.
    \item Refer to figures inline (e.g., "as demonstrated in the heart rate trend plot (Figure 1)" or "Figure 1 shows...").
    \item \textbf{Figure Captions}: In your final output, if you list figures, keep the descriptions EXTREMELY BRIEF and factual. (e.g., "Figure 1. Heart Rate vs Systolic BP."). Do not interpret the figure in the caption; interpret it in the text.
\end{itemize}

\prompthead{Global Style Constraints}
\begin{itemize}[leftmargin=*, nosep]
    \item \textbf{Tone}: Clinical, interpretive, professional.
    \item \textbf{Goal}: High-yield synthesis for decision support, not documentation.
\end{itemize}
\end{promptbox}

\begin{promptbox}[synthesizercolor]{Synthesizer Agent User Template}
\prompthead{Clinical Context}
\pyvar{\{clinical\_context\}}

\prompthead{Clinical Data}
\pyvar{\{vitals\_summary\}}

\prompthead{Calculated Metrics}
\pyvar{\{calculation\_results\}}

\prompthead{Doctor's Assessment History}
\pyvar{\{doctor\_evolution\}}

\prompthead{Visual Evidence}
Top relevant visualizations are provided below. You MUST reference these in your assessment to support your findings.
Cite them as (Figure N) in your text.
\pyvar{\{image\_reviews\}}

\prompthead{Objective}
Generate the final ``Clinical Assessment''. Ensure you populate ALL fields.
Reflect the Doctor's plan but do not just copy it --- synthesize it into a final note.
Use the provided figures to justify your acuity and diagnosis.
\end{promptbox}

\end{document}